\documentclass{article}

\usepackage{PRIMEarxiv}
\usepackage[utf8]{inputenc}
\usepackage[T1]{fontenc}
\usepackage{url}
\usepackage{booktabs}
\usepackage{amsmath}
\usepackage{amsfonts}
\usepackage{natbib}
\usepackage{microtype}
\usepackage{graphicx}
\usepackage[style=base,figurename=Fig.,labelfont=bf,labelsep=period]{caption}
\usepackage{subcaption}
\usepackage[colorlinks=true,citecolor=blue,linkcolor=black,urlcolor=blue]{hyperref}
\graphicspath{{media/}}
\begin{document}

\title{Seismic Site Response Prediction from Sparse Observations Using Finite-Element-Pretrained Latent Dynamics}

\author{
Yi Zhu \quad Su Chen\thanks{Corresponding author. E-mail address: \texttt{chensuchina@126.com}} \quad Xiaojun Li\\
State Key Laboratory of Bridge Safety and Resilience, Beijing University of Technology\\
Beijing 100124, China
}

\maketitle

\begin{abstract}
Numerical site-response predictions often deviate from observations, yet correcting these discrepancies is difficult because records are limited in both sensor coverage and number of events. This study proposes the Transfer-Enabled Forced Latent Autoencoder for Response Equations (FLARE-T) to improve these predictions by learning and calibrating low-dimensional latent dynamics that connect the base acceleration input to acceleration outputs at multiple depths. FLARE-T learns a low-dimensional response manifold and input-driven dynamics from dense finite-element simulations. It then trains a sparse encoder to map simulated sensor responses into the learned coordinates and uses limited records to calibrate the dynamics within them. A short response window initializes each prediction, while the complete base motion drives the response. The framework was evaluated using a layered-soil centrifuge test and the Lotung field vertical array. Test-set results show that FLARE-T improved multi-depth acceleration histories and $5\%$-damped pseudoacceleration response spectra relative to the original finite-element models, reducing errors at every evaluated sensor for motions of different intensities and, at Lotung, for both horizontal components. Two Lotung source models with different constitutive parameters achieved comparable test-set accuracy, indicating reduced dependence on precise prior calibration. FLARE-T therefore provides a data-efficient means of combining dense numerical response information with limited field records to improve future site-response predictions.
\end{abstract}

\keywords{Seismic site response \and Transfer learning \and Machine learning \and Reduced-order modeling}

\section{Introduction}
Seismic site response analysis is routinely used in engineering practice to estimate earthquake motions at the ground surface and at selected depths within a soil profile, providing essential inputs for the seismic design and assessment of structures, foundations, and geotechnical systems \citep{SeedEtAl1976,Borcherdt1994,BazzurroCornell2004}. For a prescribed input motion, the predicted response is governed jointly by the soil stratigraphy, shear-wave velocity profile, cyclic stress--strain behavior of the soil, and the amplitude, frequency content, and duration of the input motion \citep{RathjeEtAl2010,PuzrinEtAl1997,GriffithsEtAl2016,KimEtAl2016}. In practice, these factors cannot be characterized exactly because subsurface investigations provide limited descriptions of the site, laboratory measurements may not fully represent in-situ soil behavior, and the motions selected as input represent only a finite range of possible earthquake loading \citep{RathjeEtAl2010,GriffithsEtAl2016,HallalEtAl2022}. Additional uncertainty arises from the assumed initial state, constitutive relations, and boundary conditions used to represent the physical system \citep{Chen1985,KwokEtAl2007,RodriguezMarekEtAl2021}. Consequently, a numerical model with a reasonable physical basis does not necessarily reproduce the response of a particular site. Comparisons with vertical-array recordings have shown that predicted acceleration time histories, amplification characteristics, and response spectra can differ appreciably from their observed counterparts, with the magnitude and frequency dependence of the discrepancies varying among sites and earthquake motions \citep{YeeEtAl2013,TaoRathje2019,ZalachorisRathje2015,StewartAfshari2021,ZhuEtAl2022}. Reducing this persistent discrepancy between numerical predictions and field observations therefore remains a central practical challenge in seismic site response analysis.

Efforts to reconcile numerical predictions with observed site response have followed several complementary paths. Downhole array recordings have been used to infer in situ dynamic soil properties, calibrate constitutive parameters, and update numerical models against field measurements \citep{ChangEtAl1996,TsaiHashash2008,AssimakiEtAl2011,RotenEtAl2014}. More recently, advances in machine learning have enabled surrogate models to be trained on large ensembles of numerical results, providing efficient predictions of spectral amplification and acceleration time histories at multiple depths without repeated numerical analysis \citep{LeeEtAl2023,VanNguyenEtAl2024,IlhanEtAl2025}. Although these models can reproduce complex response patterns represented in the training simulations, agreement with numerical targets does not establish predictive accuracy against field observations because the simulated labels also reflect the assumptions and discrepancies of the underlying numerical model \citep{IlhanEtAl2025,ZhuEtAl2023}. Other studies have trained machine-learning models directly on recorded ground motions, particularly surface and borehole records from the KiK-net network, to predict site amplification, response spectra, or acceleration time histories \citep{KimEtAl2020,BergamoEtAl2021,RotenOlsen2021,ZhuEtAl2023,LiEtAl2023}. These observation-based models reduce their dependence on simulated response targets, but their development and transferability generally require datasets spanning numerous recording stations and earthquake events. Such data requirements are seldom satisfied in project-specific applications, where observations are sparse both spatially and in sample size: instruments are commonly installed at only a few depths, and the available dataset may contain only a limited number of usable events, particularly at strong shaking levels \citep{ZhuEtAl2023,ZhuEtAl2026}. Hybrid physics--data models and transfer-learning strategies have consequently been investigated as means of combining information derived from simulations, physical models, or large recording networks with limited observations from a target site \citep{TsaiHashash2008,ZhangEtAl2025,ChenEtAl2025,LiEtAl2025}. Against this background, the central problem addressed in this study is how to retain the site-specific response information contained in an existing numerical model while using limited field observations to correct its discrepancies and improve predictions for subsequent earthquake events.

Conventional numerical site-response analysis computes ground motions by solving the equations of motion together with soil constitutive relations. In the present study, the soil profile is represented as a forced input--output system, with the acceleration at the reference depth as its input and the accelerations at selected locations above that depth as its outputs. These outputs are dynamically related because they arise from the propagation of the same input motion through the soil profile. System identification studies using vertical-array records have described site response through low-order dynamical models \citep{GlaserBaise2000}, while reduced-order site models have reproduced downhole motions and surface response spectra using a limited number of generalized variables \citep{BantisEtAl2025}. These findings motivate the assumption adopted here that the dominant acceleration patterns across depth can be represented on a low-dimensional response manifold parameterized by a small set of latent variables. Each latent state corresponds to a depth-dependent acceleration pattern, and its evolution under the base motion generates the response histories. The modeling task is therefore to learn the input-driven evolution of this low-dimensional state and the mapping that reconstructs the high-dimensional acceleration outputs.

The Forced Latent Autoencoder for Response Equations (FLARE) proposed by \citet{ZhuEtAl2026FLARE} provides a framework for learning this representation. It encodes high-dimensional response histories into a small set of latent variables, identifies sparse evolution equations that explicitly incorporate external forcing, and decodes the integrated latent trajectories into observable responses. To extend this capability to data-limited site response prediction, this study proposes the Transfer-Enabled Forced Latent Autoencoder for Response Equations (FLARE-T), a transfer framework that carries the response information learned from densely sampled numerical simulations into the sparse observation space and subsequently uses limited records to correct simulation-based predictions. FLARE-T is constructed through three sequential stages. First, densely sampled acceleration responses generated by a numerical site model are used to train a source FLARE model, in which the encoder, forced latent response equations, and decoder jointly characterize the propagation of input motions through the soil profile. Second, simulated responses extracted at the instrumented depths are used to train a sparse-response encoder that maps the available sensor measurements onto the latent state learned from the dense response field, thereby establishing a consistent latent representation between the dense numerical model and the sparse sensor configuration. Third, the transferred latent dynamics and observation-domain response reconstruction are calibrated against a limited set of recorded events to correct systematic discrepancies between simulated and measured responses while retaining the response structure acquired from the source model. The framework is evaluated using multi-depth acceleration records from geotechnical centrifuge tests \citep{AfacanEtAl2014} and the field vertical array at the Lotung Large-Scale Seismic Test site \citep{BorjaEtAl1999}. Events excluded from observation-based calibration are reserved to determine whether the corrections inferred from limited records remain predictive for subsequent events, with performance evaluated using acceleration time histories and 5\%-damped response spectra at multiple depths. Two numerical source models with different degrees of prior parameter calibration are also examined in the field application to determine how the initial fidelity of the numerical model influences transfer performance. The results demonstrate that FLARE-T provides a systematic means of integrating physics-based numerical response information with sparse observations, thereby improving the predictive fidelity of site response models under data-limited conditions.

\section{Methodology}

\subsection{Problem formulation}\label{sec:problem_formulation}
For a given soil profile, let $u(t)\in\mathbb{R}$ denote the absolute horizontal acceleration prescribed at the reference depth and let $\mathbf{a}(t)\in\mathbb{R}^{m}$ denote the absolute acceleration outputs at $m$ selected locations above that depth. The numerical dataset contains dense response vectors $\mathbf{a}_{\mathrm{d}}(t)\in\mathbb{R}^{m_{\mathrm{d}}}$, whereas the recorded dataset contains sparse response vectors $\mathbf{a}_{\mathrm{s}}(t)\in\mathbb{R}^{m_{\mathrm{s}}}$ at the instrumented locations, with $m_{\mathrm{s}}\ll m_{\mathrm{d}}$. The numerical output locations and sensor depths establish the correspondence between these two response configurations.

The reduced representation uses a latent state $\mathbf{z}(t)\in\mathbb{R}^{r}$, with $r\ll m_{\mathrm{d}}$, to describe the acceleration pattern across the dense output locations. A response mapping converts this state into the physical acceleration vector, defining the low-dimensional response manifold. The latent evolution equation advances $\mathbf{z}(t)$ under the prescribed input $u(t)$, and the response mapping reconstructs the corresponding acceleration outputs. Predictions at the sensor locations are then extracted from the reconstructed dense response. Thus, the latent dimension $r$ specifies the dimension of the learned dynamical state, while $m_{\mathrm{d}}$ and $m_{\mathrm{s}}$ specify the spatial resolution of the numerical and observed outputs.

\citet{ZhuEtAl2026FLARE} introduced the Forced Latent Autoencoder for Response Equations (FLARE) to model forced systems from high-dimensional response data. FLARE combines an encoder that estimates a latent state from a response-history window, a latent dynamical model that evolves this state under a prescribed external input, and a decoder that reconstructs the physical response. For the densely sampled numerical response, the encoding and reconstruction operations are expressed as
\begin{equation}
\mathbf{z}(t)=E_T\!\left(\mathcal{H}_{L}[\mathbf{a}_{d}](t)\right),\quad \widehat{\mathbf{a}}_{d}(t)=D\!\left(\mathbf{z}(t)\right),
\label{eq:latent_representation}
\end{equation}
where $\mathcal{H}_{L}[\mathbf{a}_{d}](t)$ contains $L$ consecutive response samples ending at time $t$, $E_T$ is the source encoder, and $D$ is the decoder. The response history provides the temporal information needed to estimate the dynamic state, which cannot generally be determined from an instantaneous acceleration vector. Building on FLARE, this study proposes the Transfer-Enabled Forced Latent Autoencoder for Response Equations (FLARE-T) to connect densely sampled numerical responses with sparse measured responses. FLARE-T first learns the latent response representation and its input-driven dynamics from numerical simulations, then establishes a sparse encoder that maps responses at the instrumented locations into the same latent space, and finally uses a limited number of recorded events to correct the coefficients of the transferred dynamics. The numerical model therefore provides the spatially resolved response information and the initial dynamical representation, while the recorded data account for systematic discrepancies between simulated and measured motions. The latent variables are treated as learned coordinates of the response and are not assigned direct equivalence to soil properties, constitutive parameters, or individual vibration modes. The three stages of FLARE-T are described in Section~\ref{sec:flaret_architecture}.

After FLARE-T has been trained and adapted, prediction of a new event requires the prescribed input motion and a short initial window of measured response. The input $u(t)$ is provided over the complete analysis interval $[0,T]$ and acts as the external forcing throughout the prediction. The sparse response $\mathbf{a}_{s}(t)$ is provided only over $[0,T_{\mathrm{ini}}]$ and is used to estimate $\mathbf{z}(T_{\mathrm{ini}})$ for initialization of the first-order latent dynamics. The prediction problem is written as
\begin{equation}
\widehat{\mathbf{a}}_{s}(t)=\mathcal{P}\!\left[u(0{:}T),\mathbf{a}_{s}(0{:}T_{\mathrm{ini}})\right](t),\quad T_{\mathrm{ini}}<t\leq T,
\label{eq:prediction_problem}
\end{equation}
where $\mathcal{P}$ denotes the trained and adapted FLARE-T model. The prescribed input remains available at every integration step, whereas the measured responses are used only for initialization and are not supplied after $T_{\mathrm{ini}}$. This separation between continuous seismic input and initial response information defines the prediction setting adopted in this study and provides the basis for the offline prediction procedure presented in Section~\ref{sec:offline_prediction}.

\subsection{FLARE-T architecture}
\label{sec:flaret_architecture}
\begin{figure*}[t]
\centering
\includegraphics[width=\textwidth]{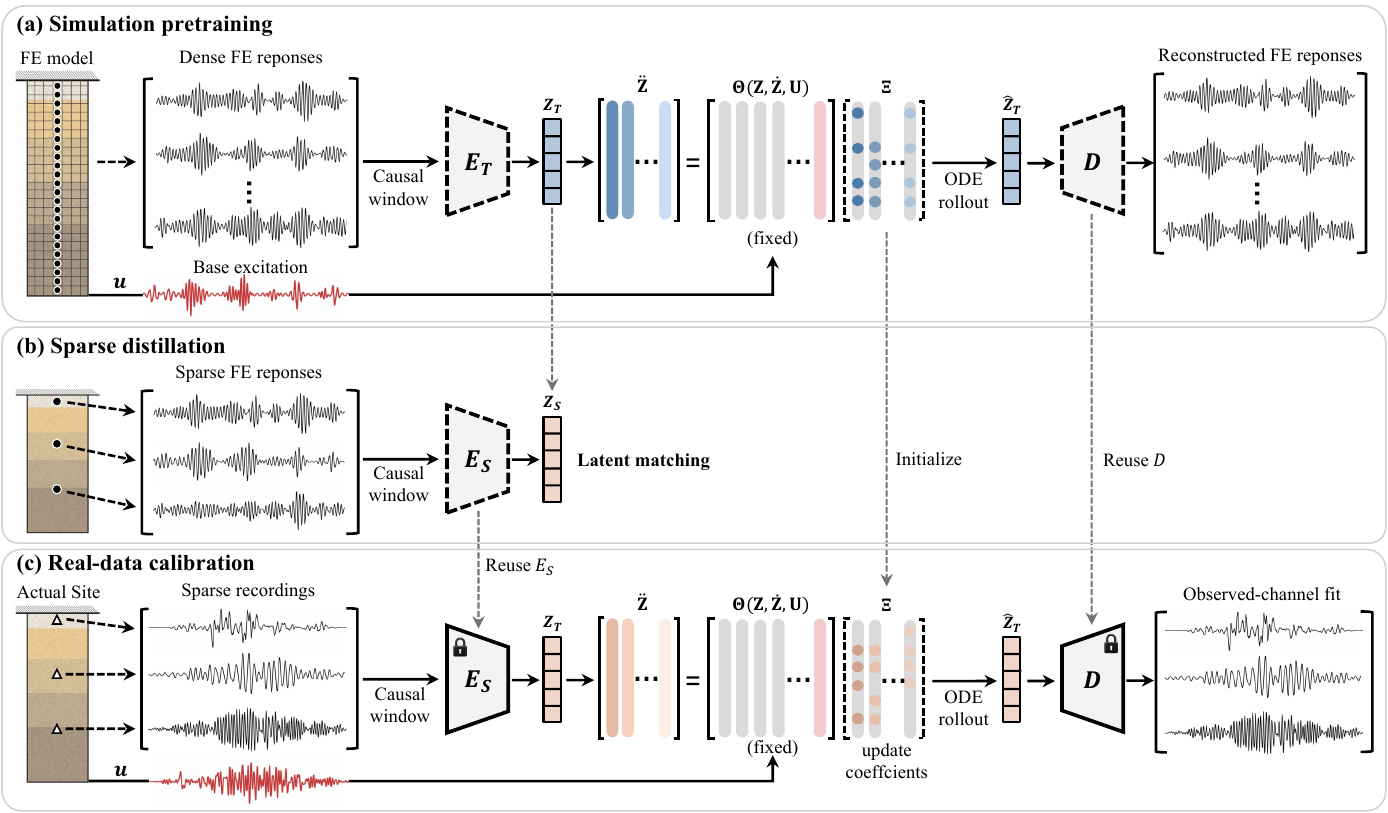}
\caption{Three-stage FLARE-T architecture: (a) simulation pretraining, (b) sparse distillation, and (c) real-data calibration.}
\label{fig:flaret_architecture}
\end{figure*}

The FLARE-T architecture comprises the three sequential stages illustrated in Fig.~\ref{fig:flaret_architecture}. In the simulation-pretraining stage, the base excitations and densely sampled finite-element responses are used to train a source encoder $E_T$, a latent response model, and a decoder $D$. In the sparse-distillation stage, simulated responses are extracted at locations corresponding to the available sensors, and a sparse encoder $E_S$ is trained to reproduce the latent states obtained from the dense response field. In the real-data-calibration stage, $E_S$ and $D$ are retained, while the coefficients of the latent response model are calibrated using a limited number of recorded events. This three-stage procedure uses numerical simulations to establish the principal response representation, sparse simulated responses to accommodate the observation configuration, and measured records to correct the latent dynamics for subsequent site-response prediction.

Stage~1 establishes the numerical source model from a collection of finite-element analyses conducted under different base motions. Each analysis is treated as a complete input--response trajectory, and the dense simulation dataset is denoted by
\begin{equation}
\mathcal{D}_{\mathrm{FE}}=\left\{\left(u_i(t),\mathbf{a}_{d,i}(t)\right)\right\}_{i=1}^{N_{\mathrm{FE}}},
\label{eq:fe_dataset}
\end{equation}
where $N_{\mathrm{FE}}$ is the number of simulated motions. Complete trajectories are assigned to the training and validation sets before response windows are extracted, ensuring that samples generated by the same base motion do not appear in different data subsets. The source model connects the dense response history, the input-driven latent dynamics, and the reconstructed response through
\begin{equation}
\mathbf{z}=E_T\!\left(\mathcal{H}_{L}[\mathbf{a}_{d}]\right),\quad \dot{\mathbf{z}}=\boldsymbol{\Theta}(\mathbf{z},u)\boldsymbol{\Xi}_{0},\quad \widehat{\mathbf{a}}_{d}=D(\mathbf{z}),
\label{eq:source_model}
\end{equation}
where $\boldsymbol{\Theta}$ is a prescribed library containing a constant term, polynomial functions of the latent coordinates, and the base-acceleration terms, and $\boldsymbol{\Xi}_{0}$ is the coefficient matrix identified from the finite-element dataset. The input motion is excluded from $E_T$ and enters only through the latent response equation. The response-history window is flattened and processed by a multilayer-perceptron encoder, and the decoder maps the latent state back to the dense acceleration response. During rollout, the first-order latent equation is integrated from the state estimated at the beginning of a training segment, and the resulting latent trajectory $\widehat{\mathbf{z}}$ is decoded without further response input. The source model is trained using
\begin{equation}
\mathcal{L}_{\mathrm{FE}}=\lambda_{\mathrm{rec}}\mathcal{L}_{\mathrm{rec}}+\lambda_{\mathrm{roll}}\mathcal{L}_{\mathrm{roll}}+\lambda_{\mathrm{lat}}\mathcal{L}_{\mathrm{lat}}+\lambda_{\mathrm{eq}}\mathcal{L}_{\mathrm{eq}}+\lambda_{1}\left\|\boldsymbol{\Xi}_{0}\right\|_{1},
\label{eq:source_loss}
\end{equation}
with
\begin{equation}
\begin{aligned}
\mathcal{L}_{\mathrm{rec}}&=\operatorname{MSE}\!\left(D(\mathbf{z}),\mathbf{a}_{d}\right), &
\mathcal{L}_{\mathrm{roll}}&=\operatorname{MSE}\!\left(D(\widehat{\mathbf{z}}),\mathbf{a}_{d}\right),\\
\mathcal{L}_{\mathrm{lat}}&=\operatorname{MSE}\!\left(\widehat{\mathbf{z}},\mathbf{z}\right), &
\mathcal{L}_{\mathrm{eq}}&=\operatorname{MSE}\!\left(\dot{\mathbf{z}},\boldsymbol{\Theta}(\mathbf{z},u)\boldsymbol{\Xi}_{0}\right).
\end{aligned}
\label{eq:source_loss_terms}
\end{equation}
Here, $\operatorname{MSE}$ denotes the mean squared error over the samples and response components included in a training segment. The encoded state is treated as a fixed target when evaluating $\mathcal{L}_{\mathrm{lat}}$, preventing this term from moving the encoder toward a degenerate representation. The reconstruction loss preserves the response information carried by the latent coordinates, the rollout losses constrain the integrated trajectory in the latent and response spaces, and the equation loss fits the local latent evolution. The $\ell_1$ penalty and sequential thresholding remove small coefficients and retain a sparse latent response equation. After convergence, $E_T$, $D$, $\boldsymbol{\Xi}_{0}$, and the retained candidate functions constitute the numerical source model.

Stage~2 transfers the response representation learned from the dense finite-element output to the sparse observation configuration. This step is required because $E_T$ expects acceleration histories from all numerical output locations and therefore cannot be applied directly to the instrumented locations. Calibrating the model immediately against measured records would combine the effects of reduced spatial observation and simulation--measurement discrepancy. FLARE-T separates these effects by first establishing the dense-to-sparse connection within the numerical domain, where synchronized dense and sparse responses are available for the same motion. A fixed observation matrix $\mathbf{C}$ maps the dense response to the sensor configuration:
\begin{equation}
\mathbf{a}_{s,\mathrm{FE}}(t)=\mathbf{C}\mathbf{a}_{d}(t).
\label{eq:sparse_observation}
\end{equation}
Each row of $\mathbf{C}$ either selects a coincident numerical output or interpolates between adjacent output locations. Dense and sparse responses are normalized separately using statistics from their respective training simulations, and the same statistics are retained for the following stage. Synchronized response-history windows are then supplied to the fixed source encoder and the sparse encoder. The latter is trained by minimizing
\begin{equation}
\mathcal{L}_{\mathrm{dist}}=\operatorname{MSE}\!\left[E_S\!\left(\mathcal{H}_{L}[\mathbf{a}_{s,\mathrm{FE}}]\right),E_T\!\left(\mathcal{H}_{L}[\mathbf{a}_{d}]\right)\right].
\label{eq:distillation_loss}
\end{equation}
Because the two response windows originate from the same finite-element trajectory and terminate at the same time, the state produced by $E_T$ provides a direct target for $E_S$. Only $E_S$ is updated during this stage; $E_T$, $D$, $\boldsymbol{\Theta}$, and $\boldsymbol{\Xi}_{0}$ remain fixed. The target states are treated as constants during optimization, and the checkpoint is selected using the latent-matching error for the numerical validation trajectories. The resulting sparse encoder estimates states in the coordinate system of the numerical source model using only the channels available from the physical sensors.

Stage~3 uses the measured events to correct the input-driven evolution of the transferred model. The sparse encoder $E_S$, decoder $D$, observation matrix $\mathbf{C}$, normalization parameters, and candidate library $\boldsymbol{\Theta}$ are fixed, and only the latent-equation coefficients are calibrated. This separation is necessary because a latent representation does not have a unique coordinate system: changing the encoder can alter the latent coordinates without representing a corresponding physical change in the response. Simultaneous adjustment of the encoder, decoder, and dynamical coefficients using a limited number of measured events could therefore reduce the fitting error by changing the coordinate system rather than by correcting the response evolution. Fixing $E_S$ and $D$ preserves the latent coordinates and the response manifold learned from the dense simulations, while fixing $\boldsymbol{\Theta}$ preserves the functional basis of the dynamics. Calibration is consequently restricted to adjusting the vector field on this fixed manifold. The numerical simulations retain their information on the depth-dependent response pattern, and the measured events correct the evolution of that pattern under the imposed motion. The calibrated coefficients are written as
\begin{equation}
\boldsymbol{\Xi}=\boldsymbol{\Xi}_{0}+\Delta\boldsymbol{\Xi},
\label{eq:coefficient_correction}
\end{equation}
where $\Delta\boldsymbol{\Xi}$ is initialized to zero and is the only trainable quantity in this stage. Calibration uses the candidate library established in Stage~1 and introduces no additional functional forms.

For each measured event, the initial latent state is estimated from the available response window. The calibrated equation is then integrated under the recorded base motion, and the decoded dense response is mapped to the instrumented locations:
\begin{equation}
\mathbf{z}_{0}=E_S\!\left(\mathcal{H}_{L}[\mathbf{a}_{s}](t_{0})\right),\quad \dot{\mathbf{z}}=\boldsymbol{\Theta}(\mathbf{z},u)\boldsymbol{\Xi},\quad \widehat{\mathbf{a}}_{s}=\mathbf{C}D(\mathbf{z}).
\label{eq:real_data_model}
\end{equation}
The response and input are processed using the normalization parameters retained from the preceding stages, and the decoded response is returned to physical units before comparison with the measurements. The measured response following initialization is used as the target of the complete rollout and is not supplied repeatedly to the encoder. The coefficient correction is determined by minimizing
\begin{equation}
\mathcal{L}_{R}=\mathcal{L}_{\mathrm{obs}}+\lambda_{\Delta}\mathcal{R}(\Delta\boldsymbol{\Xi}),
\label{eq:real_calibration_loss}
\end{equation}
where $\mathcal{L}_{\mathrm{obs}}$ is the mean squared rollout error after each sensor residual has been normalized by the corresponding standard deviation from the sparse simulation data. The regularization term $\mathcal{R}$ is calculated separately for each latent equation as the squared norm of its coefficient correction divided by the squared norm of its initial coefficient vector, and the resulting values are averaged over the latent equations. This relative penalty limits unsupported departures from the numerical dynamics while allowing equations with different coefficient magnitudes to be corrected on a comparable basis. Sequential thresholding removes small calibrated coefficients to retain a sparse model. The final coefficients are selected according to the complete-rollout error for the measured validation events, while the held-out test events are excluded from coefficient estimation and model selection.

\subsection{Offline Response Prediction}
\label{sec:offline_prediction}
After completing the three-stage model development, all components of FLARE-T are fixed for prediction, including the sparse encoder, calibrated latent dynamics, and decoder. For a new earthquake event, the horizontal input motion at the reference depth is prescribed over the complete analysis interval, while acceleration measurements at the instrumented locations are provided only within a short initial window. These initial measurements are used to establish the latent state at the start of prediction, after which the model is driven solely by the prescribed input motion. The objective of offline prediction is therefore to reproduce the subsequent acceleration responses at the selected locations through continuous latent-state evolution, without further response measurements or model-parameter updates.

Let $t_0$ denote the end of the initial observation window and the starting time of the subsequent prediction. The synchronized acceleration responses recorded at the instrumented locations over this window are processed using the same normalization parameters adopted during model development and assembled into the response history $\mathcal{H}_{L}[\mathbf{a}_{\mathrm{s}}](t_0)$. The fixed sparse encoder $E_S$ maps this response history to the initial latent state $\mathbf{z}_0$, which provides the initial condition required by the first-order latent dynamics. This initialization is expressed as
\begin{equation}
\mathbf{z}_0=\mathbf{z}(t_0)=E_S\!\left(\mathcal{H}_{L}[\mathbf{a}_{\mathrm{s}}](t_0)\right).
\label{eq:offline_initialization}
\end{equation}
\begin{figure*}[t]
\centering
\includegraphics[width=0.8\textwidth]{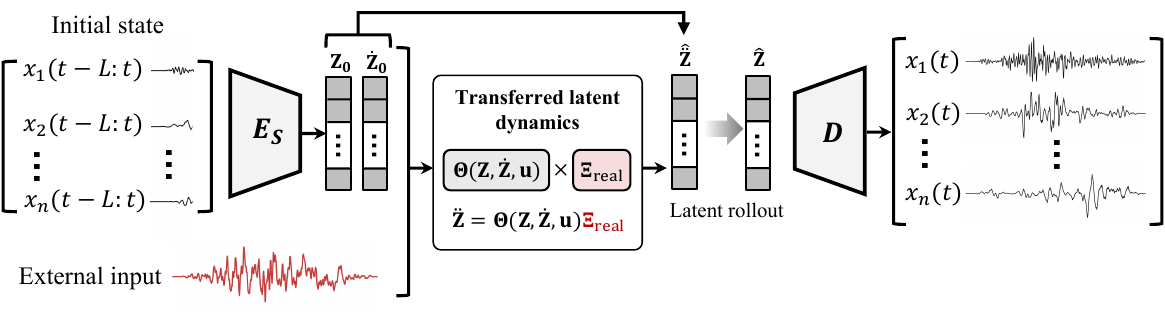}
\caption{Offline prediction using an initial response window and the prescribed input motion.}
\label{fig:offline_prediction}
\end{figure*}

Starting from $\mathbf{z}_{0}$, the calibrated latent equation is integrated over the prediction interval under the prescribed input motion. At each integration step, the current latent state and the corresponding input acceleration determine the rate of latent-state evolution, and the updated state is carried forward to the next step. The resulting latent trajectory is passed through the fixed decoder $D$ to recover the densely distributed acceleration response, after which the observation matrix $\mathbf{C}$ extracts the responses at the instrumented locations. The reconstructed responses are finally transformed back to physical units using the normalization parameters retained from model development. This rollout produces continuous acceleration histories from the end of the initialization window to the end of the event:
\begin{equation}
\begin{aligned}
\dot{\mathbf{z}}(t)&=\boldsymbol{\Theta}\!\left(\mathbf{z}(t),u(t)\right)\boldsymbol{\Xi},\quad \mathbf{z}(t_{0})=\mathbf{z}_{0},\quad t\in(t_{0},T],\\
\widehat{\mathbf{a}}_{d}(t)&=D\!\left(\mathbf{z}(t)\right),\quad \widehat{\mathbf{a}}_{s}(t)=\mathbf{C}\widehat{\mathbf{a}}_{d}(t).
\end{aligned}
\label{eq:offline_rollout}
\end{equation}

Accordingly, the initial response window and the prescribed input motion serve distinct roles in the prediction. The former determines the latent state $\mathbf{z}_{0}$ at $t_{0}$, whereas the latter drives its evolution throughout the remaining interval $(t_{0},T]$. The measured responses after $t_{0}$ are retained for comparison with the model output, and the corresponding predictions are generated as a continuous rollout of the calibrated FLARE-T model. In the following case studies, this procedure is applied to held-out events, and its predictive performance is evaluated by comparing the resulting multidepth acceleration histories and $5\%$-damped pseudo-acceleration response spectra with the corresponding measured responses.

\section{Application}
This section evaluates FLARE-T through two applications that provide distinct conditions for model development and validation. The first application uses a layered-soil centrifuge test conducted under controlled shaking, with acceleration responses recorded at multiple depths, to examine whether the response information learned from numerical simulations can be transferred to the experimental sensor configuration and corrected using a limited number of physical records \citep{AfacanEtAl2014}. The second application uses the Lotung downhole array, where earthquake records from two horizontal components allow the methodology to be evaluated under field conditions and natural seismic excitation \citep{ElgamalEtAl1995,ZeghalEtAl1995}. The two applications therefore cover model-scale and field-scale soil profiles, different material and response characteristics, and different levels of observational availability. Predictive performance is assessed using held-out records through comparisons of multidepth acceleration histories and $5\%$-damped pseudo-acceleration response spectra. The final part of this section further examines the influence of the initial numerical model by comparing FLARE-T models developed from two sets of source-model parameters for the Lotung site.

\subsection{Centrifuge Test of a Layered Soil Profile}
\label{sec:centrifuge_application}
\subsubsection{Test Configuration and Numerical Model}
\label{sec:centrifuge_model}
The physical benchmark was selected from the soft-clay centrifuge testing program conducted using the 9-m-radius geotechnical centrifuge at the University of California, Davis, and was identified as AHA02 in the experimental database \citep{AfacanEtAl2014}. The test was performed at a centrifugal acceleration of $57.2g$, for which the $49.7$-cm-high soil model represented a $28.43$-m-deep prototype profile. The profile comprised a $5.66$-m-thick upper layer of dense Monterey sand over seven layers of San Francisco Bay Mud, with adjacent clay layers separated by $0.572$-m-thick Monterey sand drainage layers. The soil model was constructed in a hinged-plate container that permitted shear deformation of the soil column, while a rigid base connected to the servo-hydraulic shaking table applied horizontal input motions. Simultaneous acceleration measurements along the depth of the model provide a controlled physical dataset for evaluating multidepth site-response prediction.

\begin{figure*}[t]
\centering
\includegraphics[width=0.7\textwidth]{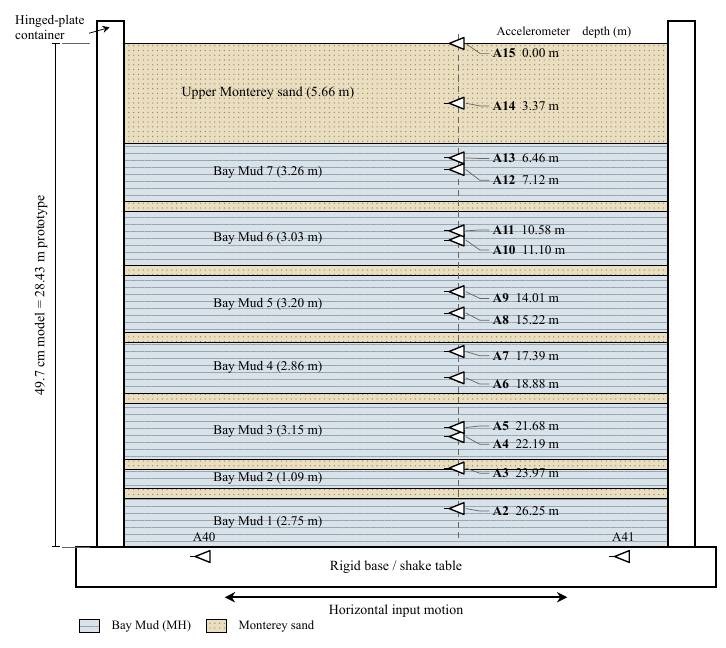}
\caption{Soil profile, accelerometer locations, and base excitation of the centrifuge model.}
\label{fig:centrifuge_profile}
\end{figure*}

Two accelerometers mounted on the rigid base, A40 and A41, recorded the imposed horizontal motion, and their average was adopted as the reference-depth input. The response vector contained the absolute accelerations recorded by the 14 accelerometers A2--A15 in the central vertical array. A2 was located $26.25$~m below the prototype ground surface, A15 was located at the surface, and the remaining sensors sampled the intervening soil layers, as illustrated in Fig.~\ref{fig:centrifuge_profile}. The input and observed response were therefore defined as
\begin{equation}
u(t)=\frac{a_{\mathrm{A40}}(t)+a_{\mathrm{A41}}(t)}{2},\quad
\mathbf{a}_{s}(t)=
\left[
a_{\mathrm{A2}}(t),a_{\mathrm{A3}}(t),\ldots,a_{\mathrm{A15}}(t)
\right]^{\mathsf{T}}.
\label{eq:centrifuge_input_output}
\end{equation}

The numerical model reproduced the prototype layer thicknesses as a $28.4284$-m-high and $1$-m-wide plane-strain soil column in Abaqus/Standard \citep{DassaultSystemes2020}. The column was discretized using 200 four-node reduced-integration plane-strain elements (CPE4R), with one element across the width and 200 elements distributed over the depth while preserving the experimental layer boundaries. The opposing nodes at each elevation were constrained to have equal horizontal displacement, and vertical displacement was restrained to maintain the shear-column kinematics; the ground surface remained traction free. A geostatic step established the initial stress state before each 30-s implicit dynamic analysis. The measured or simulated input motion was imposed as horizontal acceleration at the base, using an initial time increment of $0.001$~s and a maximum increment of $0.005$~s. Absolute horizontal accelerations were recovered at the centers of all 200 elements to form the dense numerical response. Responses at A2--A14 were obtained by linear interpolation between the adjacent numerical output locations, whereas the A15 response was taken directly from the surface nodes. Numerical and experimental trajectories were aligned to a common time origin. Dense and sparse simulated responses were normalized separately using statistics from their respective training sets, and these normalization parameters were retained when the experimental records were introduced.

The Bay Mud layers were represented using the Modified Cam--Clay model with logarithmic porous elasticity and exponential isotropic hardening \citep{RoscoeBurland1968,DassaultSystemes2020}. Let $p'$ denote the mean effective stress, $q$ the deviatoric stress, $p'_{c}$ the preconsolidation pressure, and $\varepsilon^{p}_{v}$ the plastic volumetric strain, with compression taken as positive. The yield surface and isotropic hardening relation are written as
\begin{equation}
\begin{aligned}
f_{\mathrm{MCC}}&=q^{2}+M^{2}p'\left(p'-p'_{c}\right)=0,\\
\frac{\mathrm{d}p'_{c}}{p'_{c}}&=\frac{\mathrm{d}\varepsilon^{p}_{v}}{\lambda-\kappa}.
\end{aligned}
\label{eq:centrifuge_mcc}
\end{equation}

The Monterey sand layers were represented using the Mohr--Coulomb model with linear elasticity and nonassociated plastic flow \citep{DassaultSystemes2020}. In terms of the major and minor principal effective stresses $\sigma'_1$ and $\sigma'_3$, the yield criterion is expressed as
\begin{equation}
f_{\mathrm{MC}}=
\left(\sigma'_1-\sigma'_3\right)
-\left(\sigma'_1+\sigma'_3\right)\sin\phi'
-2c'\cos\phi'=0,
\label{eq:centrifuge_mc}
\end{equation}
where $c'$ and $\phi'$ are the effective cohesion and friction angle, respectively. A dilation angle $\psi=5^{\circ}$ was adopted for the plastic potential.

The stiffness assigned to each layer was determined from its mass density $\rho_i$ and shear-wave velocity $V_{s,i}$. The corresponding shear modulus was used directly in the logarithmic elastic formulation for Bay Mud, while the Young's modulus of Monterey sand was obtained using a Poisson's ratio of $0.30$:
\begin{equation}
G_i=\rho_iV_{s,i}^{2},\quad
E_i=2\left(1+\nu_i\right)G_i.
\label{eq:centrifuge_stiffness}
\end{equation}

The initial vertical effective stress increased geostatically from zero at the ground surface to approximately $218.7$~kPa at the base. The initial horizontal effective stress in each layer was assigned using the corresponding at-rest earth-pressure coefficient:
\begin{equation}
\sigma'_{h0,i}=K_{0,i}\sigma'_{v0,i}.
\label{eq:centrifuge_initial_stress}
\end{equation}
The shear-wave velocities used in the numerical model were $95\%$ of the interpreted experimental values, and the initial preconsolidation pressures of the three upper Bay Mud layers were increased by $15\%$ during preliminary numerical-model calibration. Rayleigh coefficients $\alpha=1.1919~\mathrm{s}^{-1}$ and $\beta_{\mathrm{R}}=2.8812\times10^{-3}~\mathrm{s}$ were applied to provide $10\%$ damping at $1.0479$ and $10$~Hz. The resulting material and initial-state parameters are summarized in Table~\ref{tab:centrifuge_materials}.

\begin{table*}[t]
\centering
\caption{Material and constitutive parameters of the centrifuge finite-element model.}
\label{tab:centrifuge_materials}
{\footnotesize
\renewcommand{\arraystretch}{1.05}
\setlength{\tabcolsep}{2.8pt}
\begin{tabular*}{\textwidth}{@{\extracolsep{\fill}}lcccccccc@{}}
\toprule
\multicolumn{9}{c}{\textbf{(a) Layer-specific properties}}\\
\midrule
Layer & Depth (m) & Model & $\rho$ ($\mathrm{kg\,m^{-3}}$) & $V_s$ ($\mathrm{m\,s^{-1}}$) & $e_0$ & OCR & $K_0$ & $p'_{c0}$ (kPa)\\
\midrule
Upper Monterey sand & 0.00--5.66   & MC  & 2029 & 130.91 & --          & --   & 0.426 & --\\
Bay Mud 7           & 5.66--8.92   & MCC & 1651 & 78.85  & 1.658--1.663 & 1.27 & 0.563 & 89.56\\
Monterey sand 6     & 8.92--9.50   & MC  & 2029 & 169.67 & --          & --   & 0.426 & --\\
Bay Mud 6           & 9.50--12.53  & MCC & 1662 & 89.30  & 1.607--1.611 & 1.28 & 0.566 & 117.28\\
Monterey sand 5     & 12.53--13.10 & MC  & 2029 & 181.62 & --          & --   & 0.426 & --\\
Bay Mud 5           & 13.10--16.30 & MCC & 1672 & 102.60 & 1.570--1.573 & 1.15 & 0.536 & 142.86\\
Monterey sand 4     & 16.30--16.87 & MC  & 2029 & 192.11 & --          & --   & 0.426 & --\\
Bay Mud 4           & 16.87--19.73 & MCC & 1733 & 128.25 & 1.354--1.356 & 3.31 & 0.910 & 435.73\\
Monterey sand 3     & 19.73--20.31 & MC  & 2029 & 200.97 & --          & --   & 0.426 & --\\
Bay Mud 3           & 20.31--23.45 & MCC & 1733 & 138.70 & 1.351--1.353 & 2.76 & 0.831 & 435.73\\
Monterey sand 2     & 23.45--24.02 & MC  & 2029 & 209.36 & --          & --   & 0.426 & --\\
Bay Mud 2           & 24.02--25.11 & MCC & 1733 & 144.40 & 1.350--1.351 & 2.47 & 0.786 & 435.73\\
Monterey sand 1     & 25.11--25.68 & MC  & 2029 & 213.05 & --          & --   & 0.426 & --\\
Bay Mud 1           & 25.68--28.43 & MCC & 1733 & 148.20 & 1.348--1.349 & 2.27 & 0.753 & 435.73\\
\bottomrule
\end{tabular*}

\vspace{0.6em}

\begin{tabular*}{\textwidth}{@{\extracolsep{\fill}}lccccccccc@{}}
\toprule
\multicolumn{10}{c}{\textbf{(b) Constitutive parameters}}\\
\midrule
Material & $\lambda$ & $\kappa$ & $M$ & $\beta$ & $K$ & $\nu$ & $c'$ (kPa) & $\phi'$ ($^\circ$) & $\psi$ ($^\circ$)\\
\midrule
Bay Mud       & 0.18675 & 0.01737 & 1.20 & 1.00 & 1.00 & --   & --   & -- & --\\
Monterey sand & --      & --      & --   & --   & --   & 0.30 & 1.00 & 35 & 5\\
\bottomrule
\end{tabular*}
}
\end{table*}
\subsubsection{Dense-Response Model Training and Validation}
\label{sec:centrifuge_stage1}
The numerical dataset was generated using 100 synthetic base motions represented by nonstationary, band-limited random waveforms with varied frequency content and strong-motion duration. The target peak ground accelerations (PGAs) were stratified over the range of $0.025$--$0.55g$, which covers the approximately $0.03$--$0.54g$ range of the centrifuge records used in this study. Following baseline correction, each waveform was multiplied by a single scale factor to attain its prescribed PGA; the resulting motions had PGAs between $0.0265$ and $0.5397g$. Each motion was applied to the base of the finite-element model for a 30-s analysis, and the absolute horizontal accelerations at the 200 element-center locations were retained as the dense response field. Motions 001--080 were assigned to training and motions 081--100 to validation. This division was made at the level of complete input motions, so all response windows and sequence segments derived from a given motion remained within the same dataset.

\begin{figure*}[t]
\centering
\includegraphics[width=\textwidth]{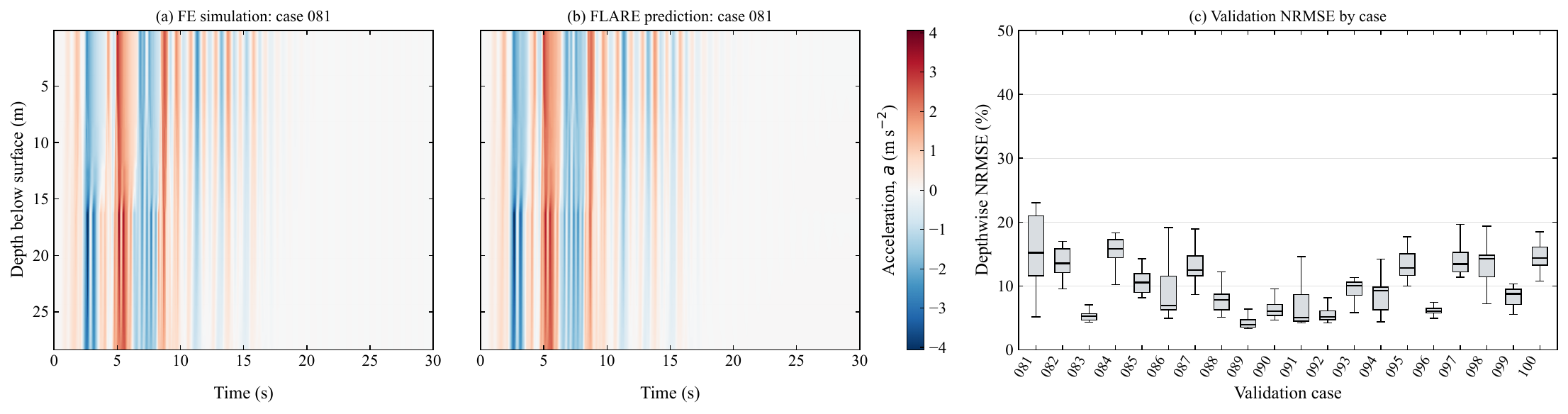}
\caption{Stage-1 validation: (a) finite-element response field for case 081; (b) corresponding model rollout; and (c) depthwise NRMSE distributions for cases 081--100.}
\label{fig:centrifuge_stage1}
\end{figure*}

For the first-stage model, each response history contained 600 time samples with an integration time step of $\Delta t=0.05$~s. The encoder used a response window of $L=3$ consecutive samples, and the 200-dimensional response field was represented by $r=8$ latent variables. The candidate library contained a constant term, all linear and quadratic terms of the latent variables, and a standalone linear term for the base acceleration; sinusoidal terms and products between the base acceleration and latent variables were excluded. For optimization, each complete response history was divided into ten contiguous 60-sample segments, resulting in 800 training sequences and 200 validation sequences, while model selection was performed using the complete validation motions. The retained model minimized the validation score formed from the rollout, reconstruction, and latent-rollout errors with relative weights of $1$, $0.25$, and $0.05$, respectively. Prediction accuracy at the $j$th response location was measured using the normalized root-mean-square error (NRMSE),

\begin{equation}
\mathrm{NRMSE}_{j}
=
\frac{\operatorname{RMS}\!\left[\widehat{a}_{j}(t)-a_{j}(t)\right]}
{\operatorname{RMS}\!\left[a_{j}(t)\right]}
\times 100\% ,
\label{eq:nrmse}
\end{equation}
where $a_j(t)$ and $\widehat{a}_j(t)$ are the finite-element response and model prediction, respectively. Each box in Fig.~\ref{fig:centrifuge_stage1} represents the distribution of $\mathrm{NRMSE}_{j}$ over the 200 response depths for one complete validation motion.

\begin{table}[htbp]
\caption{Data and model settings for the centrifuge application.}
\label{tab:aha02_settings}
\centering
\small
\renewcommand{\arraystretch}{1.15}
\begin{tabular*}{\textwidth}{@{\extracolsep{\fill}}p{0.17\textwidth}p{0.21\textwidth}p{0.23\textwidth}p{0.30\textwidth}@{}}
\hline
& Stage 1 & Stage 2 & Stage 3 \\
\hline
Data source
& Dense FE responses
& Dense--sparse FE pairs
& Centrifuge records \\
Training
& Cases 001--080
& Cases 001--080
& S006--S008, S010--S012, S015--S017, S019--S021 \\
Validation
& Cases 081--100
& Cases 081--100
& S013, S018 \\
Test
& --
& --
& S009, S014, S022 \\
Input
& $\mathcal{H}_{L}[\mathbf{a}_{\mathrm{d}}],\,u$
& $\mathcal{H}_{L}[\mathbf{a}_{\mathrm{s,FE}}]$
& $\mathcal{H}_{L}[\mathbf{a}_{\mathrm{s}}],\,u$ \\
Target
& $\mathbf{a}_{\mathrm{d}}$
& $\mathbf{z}$
& $\mathbf{a}_{\mathrm{s}}$ \\
Channels
& 200
& 14
& 14 \\
$L$
& 3
& 3
& 3 \\
$r$
& 8
& 8
& 8 \\
\hline
\end{tabular*}
\end{table}

Figure~\ref{fig:centrifuge_stage1}(a) and Fig.~\ref{fig:centrifuge_stage1}(b) compare the finite-element response field and the corresponding full-record rollout for validation motion 081. The model reproduces the timing and depth-dependent propagation of the principal response bands, together with their amplitude distribution and decay after the strong-motion interval. The remaining discrepancies are concentrated near the largest response pulses and vary gradually with depth, without altering the overall response pattern. For the 20 validation motions in Fig.~\ref{fig:centrifuge_stage1}(c), the median depthwise NRMSE ranges from $3.9\%$ to $15.8\%$, and the median over all motion-depth combinations is $9.6\%$. The relatively compact interquartile ranges for most motions indicate that the prediction accuracy is generally maintained throughout the profile, providing the dense numerical response model required for the subsequent transfer to the experimental sensor configuration.

\subsubsection{Sparse-Response Transfer and Validation}
\label{sec:centrifuge_stage2}
The source encoder developed in Stage~1 estimates the latent state from acceleration histories at all 200 finite-element output locations, whereas the centrifuge measurements are available only at accelerometers A2--A15. The source encoder therefore cannot be applied directly to the experimental records. To reconcile these different observation configurations before introducing measured data, acceleration responses at A2--A15 were extracted from each finite-element simulation to form a sparse dataset consistent with the centrifuge instrumentation. These sparse responses were used to train the sparse encoder $E_S$ to recover the latent states produced by the source encoder $E_T$. This stage establishes a mapping from the instrumented responses to the latent coordinates learned from the dense numerical response field.

For each simulated motion, a response-history window of length $L=3$ was formed from the 14 acceleration channels corresponding to A2--A15 and supplied to $E_S$. The synchronized dense-response window was supplied to the fixed source encoder to provide the target latent state. Cases 001--080 were used for training and cases 081--100 were retained for validation, following the complete-motion allocation summarized in Table~\ref{tab:aha02_settings}. Only the parameters of $E_S$ were updated by minimizing the latent-state matching loss in Eq.~\eqref{eq:distillation_loss}; the source encoder $E_T$, decoder $D$, candidate library $\boldsymbol{\Theta}$, and coefficient matrix $\boldsymbol{\Xi}_{0}$ remained fixed. The final sparse encoder was selected according to the lowest latent-state matching error over the validation motions.

\begin{figure*}[t]
\centering
\includegraphics[width=\textwidth]{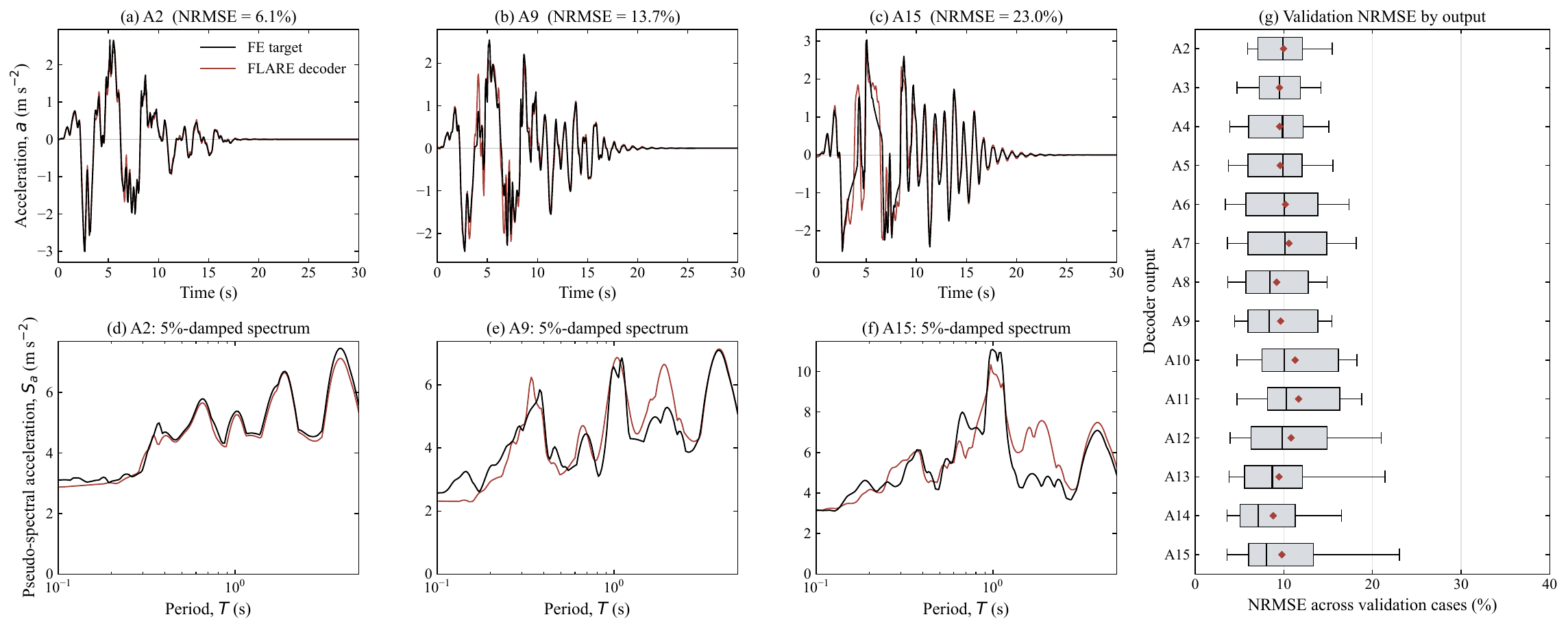}
\caption{Stage-2 validation: (a)--(c) reconstructed acceleration time histories at A2, A9, and A15; (d)--(f) corresponding $5\%$-damped pseudoacceleration response spectra; and (g) NRMSE distributions at A2--A15.}
\label{fig:aha02_stage2}
\end{figure*}

Figure~\ref{fig:aha02_stage2} evaluates the transferred representation for a representative validation motion and for the complete validation set. Accelerometers A2, A9, and A15 represent response locations near the base, at an intermediate depth, and at the ground surface, respectively. As shown in Figs.~\ref{fig:aha02_stage2}(a)--\ref{fig:aha02_stage2}(c), the responses reconstructed through the sparse encoder and fixed decoder reproduce the principal phases, amplitudes, and duration of the finite-element targets. The agreement is closest at A2 and decreases toward A15, with NRMSE values of $6.1\%$, $13.7\%$, and $23.0\%$ for the three illustrated responses. The corresponding $5\%$-damped pseudoacceleration response spectra in Figs.~\ref{fig:aha02_stage2}(d)--\ref{fig:aha02_stage2}(f) preserve the principal spectral peaks, although local amplitude differences become more evident at A9 and A15. In Fig.~\ref{fig:aha02_stage2}(g), each boxplot represents the NRMSE distribution for one accelerometer across validation cases 081--100. The distributions remain centered near $10\%$ at most locations, while several upper sensors exhibit a wider range of errors among the validation motions. These results show that the sparse encoder can locate the response state within the existing latent coordinates using only the channels available from the centrifuge instrumentation.

\subsubsection{Record-Based Calibration and Held-Out Prediction}
\label{sec:centrifuge_stage3}
Stage~3 used the measured centrifuge motions according to the allocation summarized in Table~\ref{tab:aha02_settings}. Records S006--S008, S010--S012, S015--S017, and S019--S021 were used to calibrate the latent-dynamics coefficients, while S013 and S018 were used exclusively for model selection based on the validation rollout error. Records S009, S014, and S022 were excluded from both coefficient calibration and model selection and were retained for final testing. Throughout this stage, the sparse encoder $E_S$, decoder $D$, candidate library $\boldsymbol{\Theta}$, and response mapping $\mathbf{C}$ remained fixed. Only the coefficient matrix governing the latent-state evolution was adjusted using the calibration records.

\begin{figure*}[t]
\centering
\includegraphics[width=0.92\textwidth]{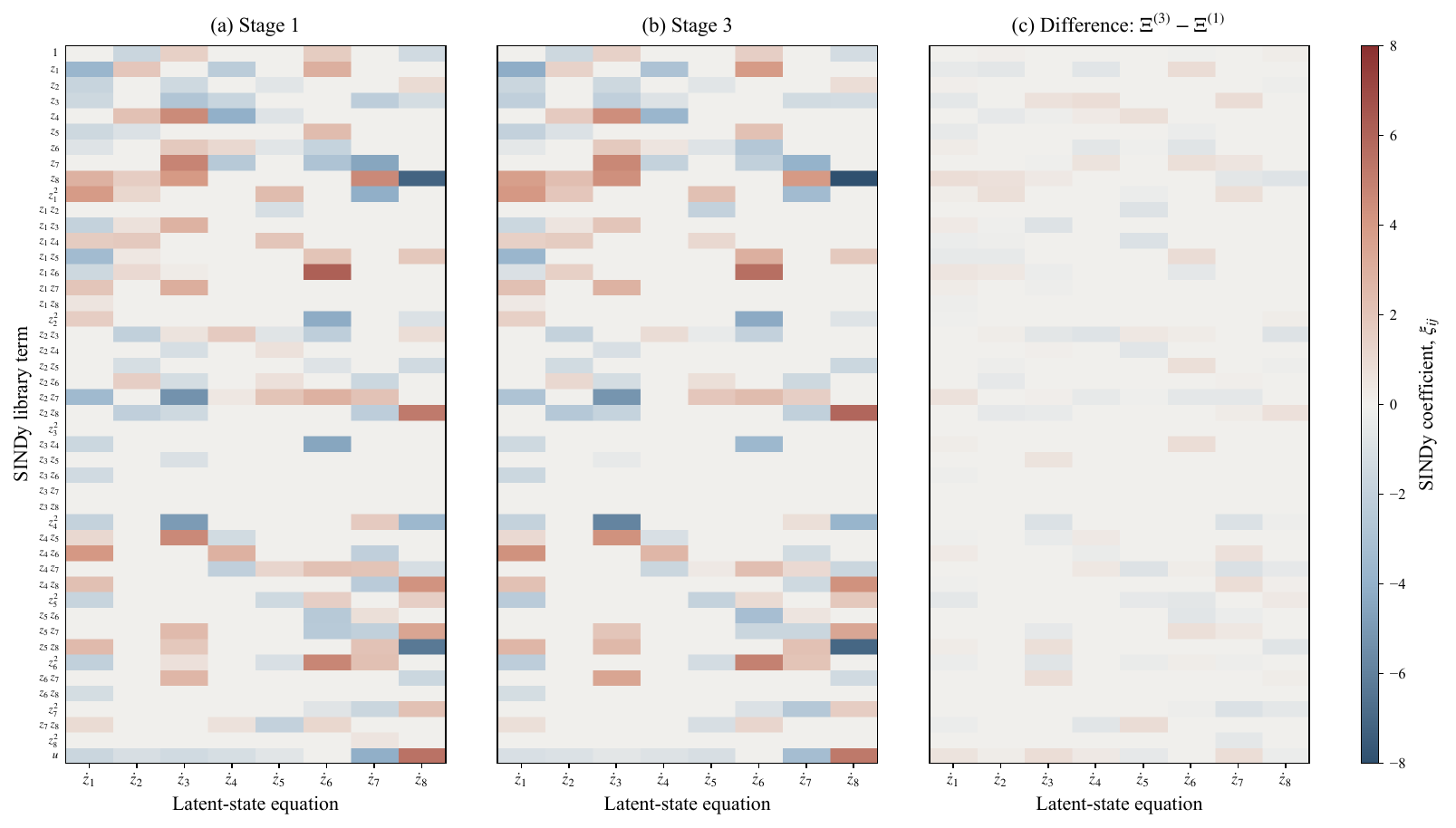}
\caption{Latent-dynamics coefficients: (a) Stage~1, (b) Stage~3, and (c) their difference.}
\label{fig:aha02_coefficients}
\end{figure*}

Because Stage~2 did not modify the latent-dynamics coefficients, the Stage~1 coefficient matrix in Fig.~\ref{fig:aha02_coefficients}(a) also represents the model immediately before record-based calibration. The calibrated coefficients in Fig.~\ref{fig:aha02_coefficients}(b) retain the dominant positive and negative patterns of the source model, while the difference in Fig.~\ref{fig:aha02_coefficients}(c) shows localized changes of smaller magnitude. The cosine similarity between the two coefficient matrices is $0.98$, confirming that calibration preserved their overall structure while adjusting selected terms. Since the sparse encoder and decoder remained fixed, these coefficient changes represent corrections to the latent-state evolution within the response coordinates established from the finite-element simulations.

\begin{figure*}[t]
\centering
\includegraphics[width=\textwidth]{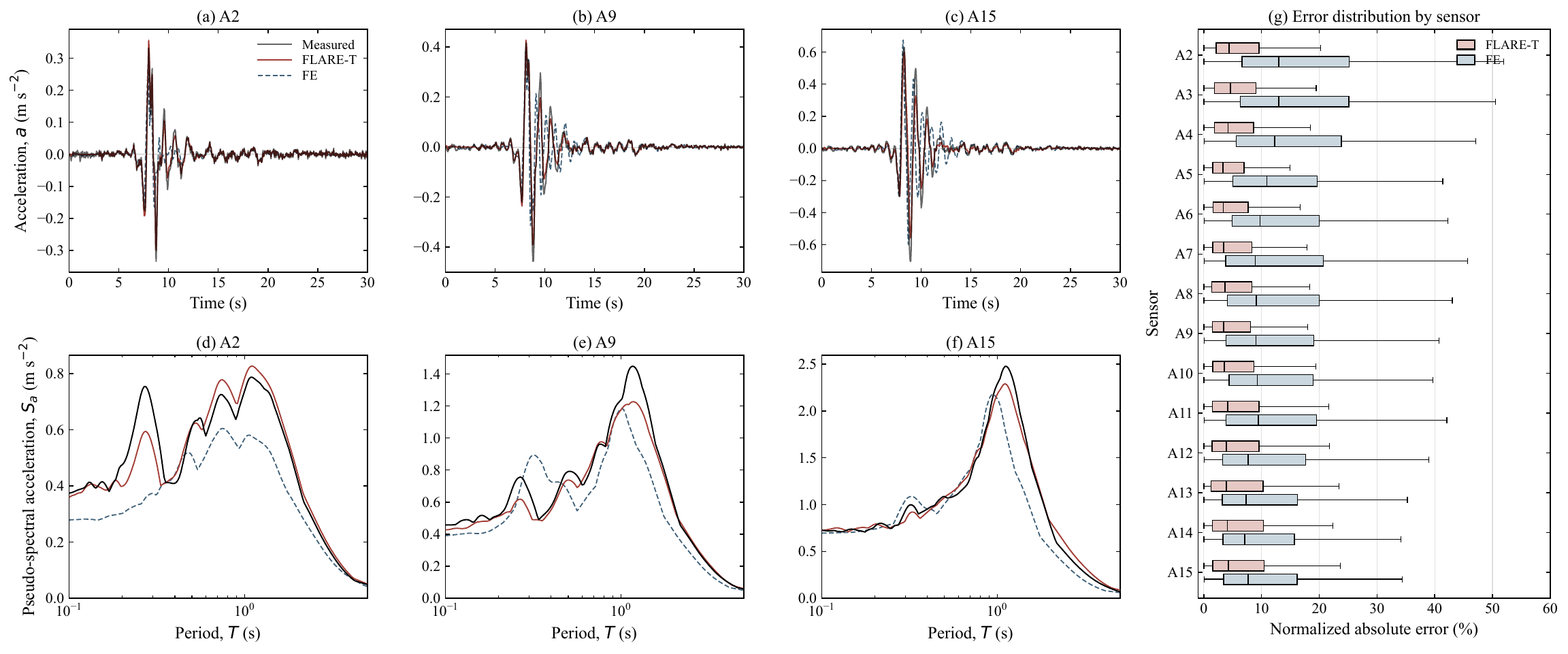}
\caption{Held-out motion S009: (a)--(c) acceleration time histories; (d)--(f) $5\%$-damped pseudoacceleration response spectra; and (g) sensorwise error distributions.}
\label{fig:aha02_s009}
\end{figure*}

\begin{figure*}[t]
\centering
\includegraphics[width=\textwidth]{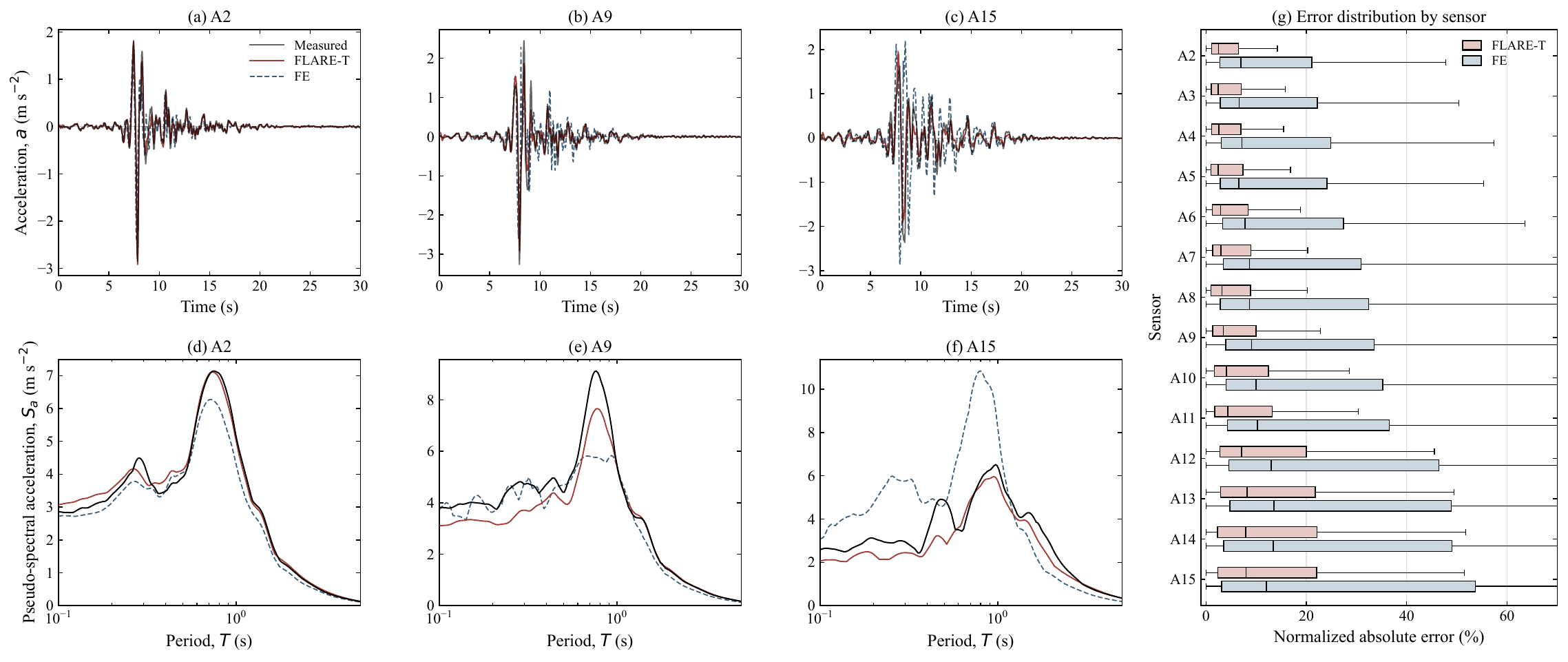}
\caption{Held-out motion S014: (a)--(c) acceleration time histories; (d)--(f) $5\%$-damped pseudoacceleration response spectra; and (g) sensorwise error distributions.}
\label{fig:aha02_s014}
\end{figure*}

\begin{figure*}[t]
\centering
\includegraphics[width=\textwidth]{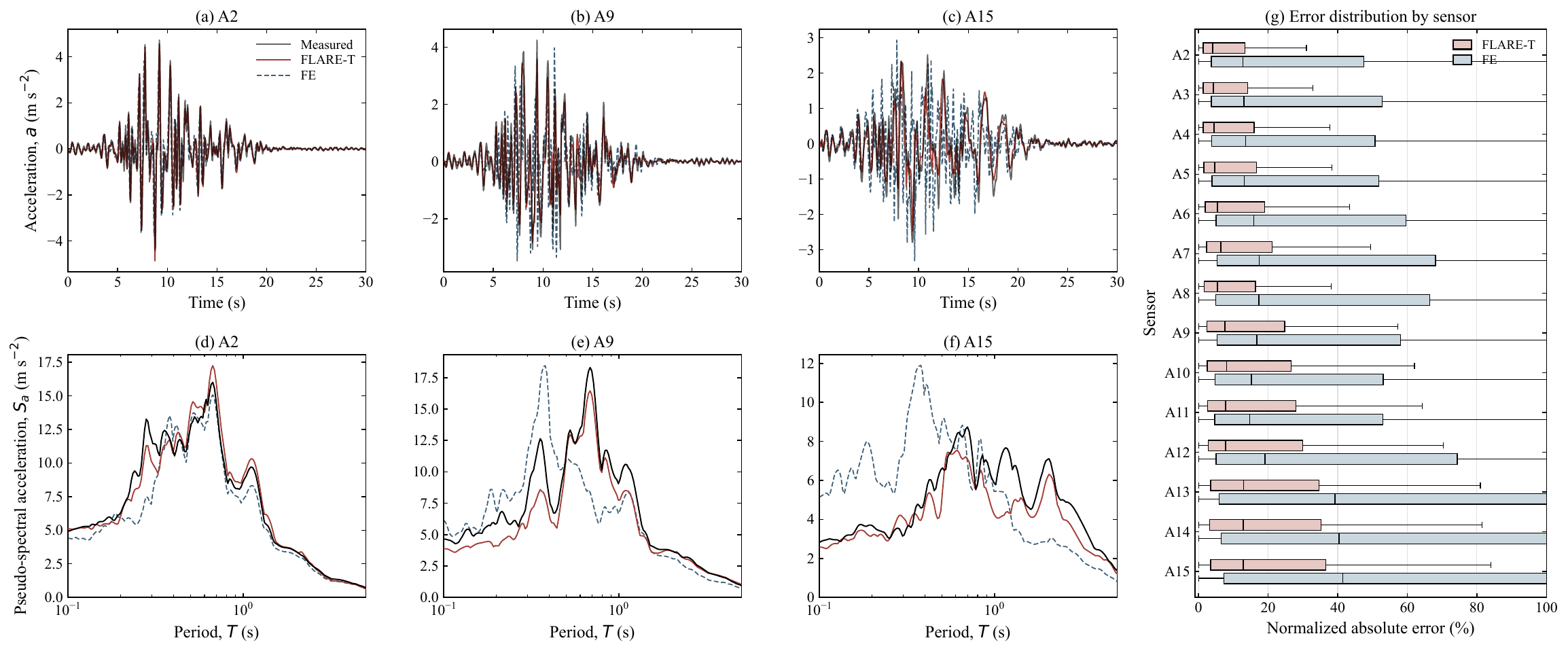}
\caption{Held-out motion S022: (a)--(c) acceleration time histories; (d)--(f) $5\%$-damped pseudoacceleration response spectra; and (g) sensorwise error distributions.}
\label{fig:aha02_s022}
\end{figure*}

Each held-out motion was initialized using the same three-sample response window from A2--A15. The latent equation was subsequently integrated over the remaining analysis interval under the complete measured base motion, and no additional response measurements were supplied during prediction. The resulting FLARE-T responses were compared with the centrifuge measurements and with the finite-element responses obtained without record-based correction. For a prediction $a_{j}^{\mathrm{p}}(t)$ at sensor $j$, the time-dependent normalized absolute error was defined as

\begin{equation}
e_{j}(t)
=
\frac{\left|a_{j}^{\mathrm{p}}(t)-a_{j}(t)\right|}
{\operatorname{RMS}\!\left[a_{j}(t)\right]}
\times 100\% ,
\label{eq:timewise_error}
\end{equation}

where $a_{j}(t)$ is the measured response and $a_{j}^{\mathrm{p}}(t)$ denotes either the FLARE-T or finite-element result. The pseudoacceleration response spectra were calculated with $5\%$ damping over periods from $0.10$ to $5.00$~s. In Figs.~\ref{fig:aha02_s009}(g)--\ref{fig:aha02_s022}(g), each boxplot contains the values of $e_j(t)$ over all time samples of one test motion at the indicated sensor.

For the lower-amplitude motion S009, FLARE-T closely follows the measured phase and principal acceleration cycles at A2, A9, and A15 and reproduces the decay of the response following the main shaking interval. The finite-element result shows increasingly evident phase and amplitude differences toward A9 and A15, whereas FLARE-T remains aligned with the measured oscillations. The principal spectral peaks at A9 and A15 are also represented more closely by FLARE-T, although the shorter-period peak at A2 remains underestimated. The sensorwise distributions in Fig.~\ref{fig:aha02_s009}(g) show lower median errors and narrower interquartile ranges for FLARE-T than for the finite-element model throughout A2--A15.

For the higher-amplitude motion S014, both predictions reproduce the principal acceleration pulse near the base, but their differences become more pronounced at the intermediate and surface locations. FLARE-T more closely captures the phase, peak sequence, and subsequent decay at A9 and A15. Its response spectra reproduce the dominant A2 and A9 peaks more accurately and substantially reduce the excessive spectral amplification predicted by the finite-element model at A15. Consistent reductions in the median and spread of the time-dependent error are observed at every sensor in Fig.~\ref{fig:aha02_s014}(g), indicating that the improvement is maintained as the shaking amplitude increases.

Motion S022 produces the largest response amplitudes among the three held-out events. FLARE-T continues to reproduce the timing of the principal wave groups and the decay of the measured motions more closely than the finite-element model at the three representative locations. The agreement is strongest at A2, while residual amplitude differences remain at A9 and A15 during the most intense portion of the response. FLARE-T provides a closer representation of the measured spectral shape and reduces the pronounced short-period overprediction of the finite-element model at the surface, although differences in individual spectral peaks remain. Figure~\ref{fig:aha02_s022}(g) nevertheless shows consistently lower error distributions for FLARE-T at all sensors, demonstrating that the record-based correction remains effective for this stronger event.

Across the three held-out motions, FLARE-T provides more consistent multi-depth acceleration histories than the finite-element model and generally improves the amplitudes and locations of the principal response-spectrum peaks. The reductions in the sensorwise error distributions are maintained over the range of shaking amplitudes represented by the test events, with the largest improvements occurring at locations where discrepancies in the original finite-element responses accumulate during upward propagation. The remaining spectral differences for the strongest motion indicate that the correction does not reproduce every local peak, but the overall results show that the information obtained from the calibration records remains predictive for events excluded from model development. The following application evaluates the same framework using field vertical-array observations and separately modeled horizontal components.

\subsection{Field Vertical-Array Records at the Lotung Site}
\subsubsection{Site Conditions and Numerical Model}
\label{sec:lotung_model}

The Lotung Large-Scale Seismic Test (LSST) site in northeastern Taiwan was established for field investigations of soil--structure interaction and was instrumented with free-field surface and downhole accelerometers. Because these instruments recorded the same earthquakes at several elevations within the soil deposit, the resulting vertical-array data have been widely used to identify site dynamic properties, infer depth-dependent stiffness and nonlinear soil behavior, and evaluate numerical site-response models \citep{ElgamalEtAl1995,ZeghalEtAl1995,ChangEtAl1996,GlaserBaise2000,BorjaEtAl1999,LeeEtAl2006Lotung}. The present application uses the free-field records to evaluate whether FLARE-T can predict depth-dependent motions for earthquake events excluded from record-based calibration.

\begin{figure}[!t]
\centering
\includegraphics[width=0.6\textwidth]{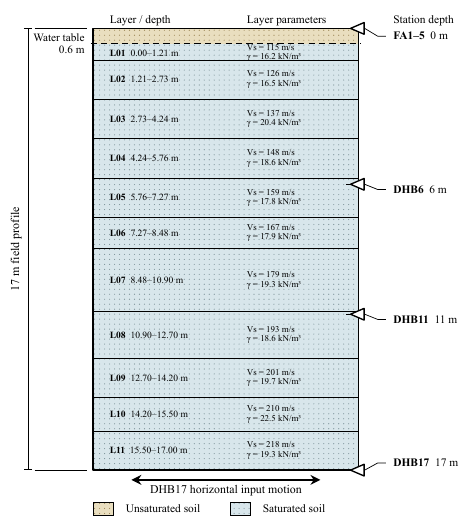}
\caption{Lotung soil profile and free-field accelerometers used for model input and response prediction.}
\label{fig:lotung_site}
\end{figure}

The modeled profile extends from the ground surface to a depth of 17~m and predominantly consists of layered silty sand and sandy silt deposits. The groundwater table is located 0.6~m below the ground surface. The horizontal acceleration recorded by DHB17 at a depth of 17~m is prescribed as the input motion, while the records from DHB11 at 11~m, DHB6 at 6~m, and FA1-5 at the ground surface define the response quantities considered in this study. These stations provide synchronized acceleration histories for multiple earthquake events in both the east--west and north--south directions, allowing the prediction performance to be examined for separate horizontal components under field conditions. The soil profile and the accelerometer locations used in the analysis are shown in Fig.~\ref{fig:lotung_site}.

The numerical model was developed in Abaqus/Standard as a 17-m-deep and 0.085-m-wide plane-strain soil column. The profile was discretized using 200 four-node reduced-integration elements (CPE4R), giving 201 nodal elevations and 402 nodes. Nodes at the same elevation on the two lateral boundaries were constrained to undergo identical motion, the ground surface was traction free, and vertical displacement was restrained at the base. The prescribed horizontal acceleration was applied uniformly along the base. The initial effective-stress field was established under self-weight using $K_{0}=0.50$, with the groundwater condition introduced at a depth of 0.6~m. Each dynamic analysis covered 15~s, with an initial time increment of 0.001~s, a maximum increment of 0.005~s, and an output interval of 0.005~s. Absolute accelerations were obtained at 200 elevations above the input boundary to form the dense numerical response field; responses corresponding to DHB11, DHB6, and FA1-5 were subsequently extracted to form the sparse simulated dataset.

The soil layers were represented using the Dafalias--Manzari SANISAND model, which describes pressure-dependent stiffness, state-dependent hardening and dilatancy, and the effect of fabric evolution under cyclic loading \citep{DafaliasManzari2004}. The elastic shear and bulk moduli are expressed as

\begin{equation}
\begin{aligned}
G &= G_{0}\,p_{a}\frac{(2.97-e)^{2}}{1+e}\left(\frac{\bar{p}}{p_{a}}\right)^{1/2},\\
K &= \frac{2(1+\nu)}{3(1-2\nu)}G ,
\end{aligned}
\label{eq:lotung_elastic_moduli}
\end{equation}
where $e$ is the current void ratio, $p_{a}$ is atmospheric pressure, $\nu$ is Poisson's ratio, and $\bar{p}$ is the mean effective stress used in the constitutive integration. A layer-specific value of $G_{0}$ was determined so that the initial shear modulus $G_{\max}=\rho V_{s}^{2}$ reproduced the shear-wave velocity assigned to each layer in Fig.~\ref{fig:lotung_site}. The dependence of the constitutive response on the current density and confinement is introduced through the state parameter

\begin{equation}
\Psi=e-e_{c},\quad e_{c}=e_{0}-\lambda_{c}\left(\frac{\bar{p}}{p_{a}}\right)^{\xi},
\label{eq:lotung_state_parameter}
\end{equation}
where $e_{c}$ is the critical-state void ratio at the current mean effective stress. Plastic loading is governed by the yield condition

\begin{equation}
f=\left\|\mathbf{s}-\bar{p}\boldsymbol{\alpha}\right\|-\sqrt{\frac{2}{3}}\,m\bar{p}=0,
\label{eq:lotung_sanisand_yield}
\end{equation}
where $\mathbf{s}$ is the deviatoric effective-stress tensor, $\boldsymbol{\alpha}$ is the back-stress-ratio tensor, and $m$ controls the opening of the yield surface. The evolution of $\boldsymbol{\alpha}$ and the fabric tensor governs the hardening, dilatancy, and response following load reversal. Below the groundwater table, the excess pore-pressure response was evaluated within the constitutive implementation using a water bulk modulus of 2.2~GPa. The parameter set used for the numerical model is summarized in Table~\ref{tab:lotung_sanisand_parameters}.

\begin{table}[!t]
\caption{SANISAND parameters used for the Lotung finite-element model.}
\label{tab:lotung_sanisand_parameters}
\centering
\small
\begin{tabular*}{\textwidth}{@{\extracolsep{\fill}}l c l c}
\hline
Parameter & Value & Parameter & Value \\
\hline
$p_{a}$ (kPa) & 100 & $e_{0}$ & 0.934 \\
$\lambda_{c}$ & 0.019 & $\xi$ & 0.70 \\
$M_{c}$ & 1.25 & $M_{e}$ & 0.882 \\
$m$ & 0.0085 & $G_{0}$ & 266.99--430.82 \\
$\nu$ & 0.05 & $e_{\mathrm{ini}}$ & 0.86 \\
$h_{0}$ & 6.00 & $c_{h}$ & 0.968 \\
$n_{b}$ & 1.10 & $A_{0}$ & 0.704 \\
$n_{d}$ & 3.50 & $z_{\max}$ & 4.00 \\
$c_{z}$ & 600 & $p_{t}/p_{a}$ & 0.05 \\
$K_{w}$ (GPa) & 2.20 & $K_{0}$ & 0.50 \\
\hline
\end{tabular*}
\end{table}

\subsubsection{Dense-Response Model Training and Validation}
\label{sec:lotung_stage1}

A dataset of 100 nonstationary horizontal input motions was used to generate the numerical responses for source-model training. The motions had peak ground accelerations ranging from $0.0075$ to $0.112g$ and comprised 60 weak motions with $\mathrm{PGA}<0.02g$, 20 moderate motions with $0.02g\leq\mathrm{PGA}<0.06g$, and 20 strong motions with $\mathrm{PGA}\geq0.06g$. Each motion had a duration of 15~s and was applied at the base of the finite-element model described in Section~\ref{sec:lotung_model}. The simulated motions represent a scalar horizontal component and were not assigned east--west or north--south labels; consequently, the same numerical dataset was used to establish the common source model for both horizontal components. Cases~001--080 were assigned to training and Cases~081--100 were reserved for validation, with each complete input motion and its corresponding response field retained within a single data partition.

Of the 200 numerical output locations, 100 approximately uniformly spaced response channels were retained for Stage~1. The source encoder used a response-window length of $L=3$, and the depth-distributed acceleration response was represented by $r=14$ latent variables. The candidate library contained constant, linear, and quadratic functions of the latent state, while the base acceleration entered the latent dynamics as a separate linear forcing term. The synchronized input and response histories were supplied to the model at a time interval of 0.05~s. After allocation of the complete motions to the training and validation sets, each motion was divided into five contiguous sequences for optimization. Model selection was based on the error obtained by rolling the latent dynamics through the complete validation motions and reconstructing all 100 response channels. Prediction errors were evaluated using the NRMSE defined in Eq.~\eqref{eq:nrmse}. Table~\ref{tab:lotung_model_settings} summarizes the data allocation and principal model settings for all three stages of the Lotung application.

\begin{table}[!t]
\caption{Data and model settings for the Lotung application.}
\label{tab:lotung_model_settings}
\centering
\small
\renewcommand{\arraystretch}{1.15}
\begin{tabular*}{\textwidth}{@{\extracolsep{\fill}}llll@{}}
\hline
& Stage 1 & Stage 2 & Stage 3 \\
\hline
Data source
& Dense FE responses
& Dense--sparse FE pairs
& Vertical-array records \\
Training
& Cases 001--080
& Cases 001--080
& \shortstack[l]{LSST05--06, LSST08--10,\\LSST12--13, LSST18} \\
Validation
& Cases 081--100
& Cases 081--100
& LSST11, LSST15--16 \\
Test
& --
& --
& LSST07, LSST14, LSST17 \\
Input
& $\mathcal{H}_{L}[\mathbf{a}_{\mathrm{d}}],\,u$
& $\mathcal{H}_{L}[\mathbf{a}_{\mathrm{s,FE}}]$
& $\mathcal{H}_{L}[\mathbf{a}_{\mathrm{s}}],\,u$ \\
Target
& $\mathbf{a}_{\mathrm{d}}$
& $\mathbf{z}$
& $\mathbf{a}_{\mathrm{s}}$ \\
Channels
& 100
& 3
& 3 \\
$L$
& 3
& 5
& 5 \\
$r$
& 14
& 14
& 14 \\
Components
& Common to EW/NS
& Common to EW/NS
& EW/NS separate \\
\hline
\end{tabular*}
\end{table}

\begin{figure}[!t]
\centering
\includegraphics[width=\textwidth]{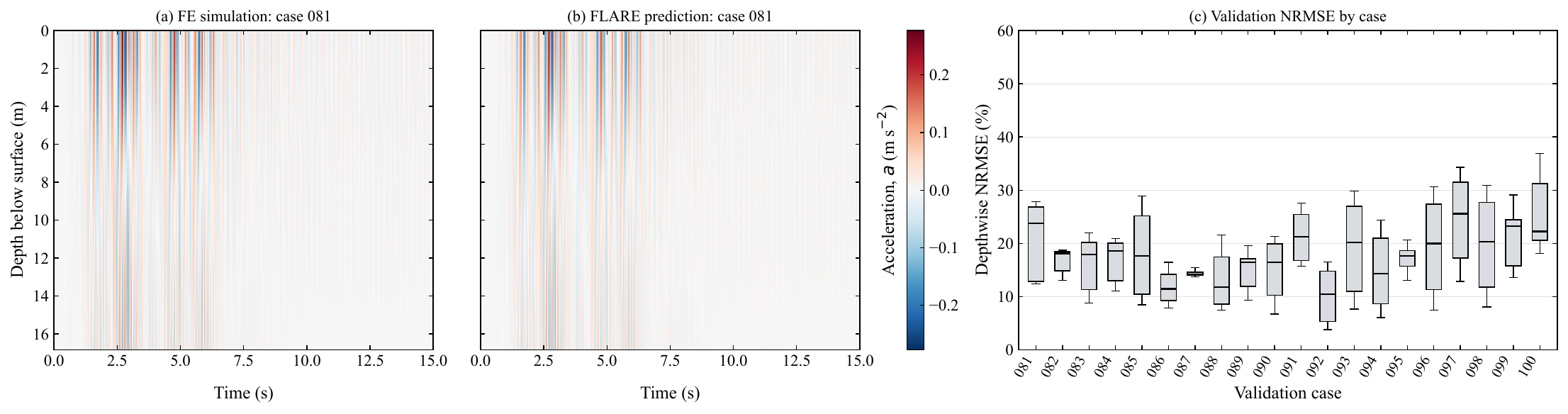}
\caption{Stage~1 validation: (a) finite-element response field for Case~081, (b) complete rollout prediction, and (c) depthwise NRMSE for Cases~081--100.}
\label{fig:lotung_stage1}
\end{figure}

Figures~\ref{fig:lotung_stage1}(a) and \ref{fig:lotung_stage1}(b) compare the finite-element response field and the complete rollout prediction for validation Case~081. The prediction reproduces the timing and depth coherence of the principal acceleration cycles, including the concentration of stronger response during the main shaking interval and the subsequent decay in amplitude. The major positive and negative acceleration bands are recovered throughout the modeled depth, although localized differences remain in the amplitudes of several strong cycles. Figure~\ref{fig:lotung_stage1}(c) extends the evaluation to Cases~081--100. Each boxplot contains the NRMSE values obtained at the 100 response depths for one validation motion. The median depthwise errors are approximately 10--26\%, while differences in the widths of the distributions show that the variation of prediction accuracy with depth depends on the input motion. These results confirm that the source model retains the principal temporal and depth-dependent characteristics of numerical site response for motions excluded from training.

\subsubsection{Sparse-Response Transfer and Validation}
\label{sec:lotung_stage2}

The Stage~1 encoder requires acceleration histories from 100 depth-distributed response channels, whereas the Lotung vertical array provides responses only at DHB11, DHB6, and FA1-5 within the modeled profile. Responses at these three locations were therefore extracted from each finite-element response field to form a sparse simulated dataset with the same observation configuration as the field records. This paired dataset was used to train a sparse encoder that estimates the latent coordinates established in Stage~1 from the three available response histories.

The sparse encoder received acceleration windows from DHB11, DHB6, and FA1-5, using a window length of $L=5$ and the previously established latent dimension of $r=14$. Cases~001--080 were used for training, and Cases~081--100 were used for validation, consistent with the simulation partition adopted in Stage~1. Only the parameters of the sparse encoder were updated during this stage; the source encoder, decoder, latent dynamics, candidate library, response mapping, and normalization parameters remained fixed. Training employed the latent-state matching loss defined in Eq.~\eqref{eq:distillation_loss}, and the model with the lowest latent-state matching error over the validation cases was retained. The corresponding data allocation and model settings are summarized in Table~\ref{tab:lotung_model_settings}.

\begin{figure}[!t]
\centering
\includegraphics[width=\textwidth]{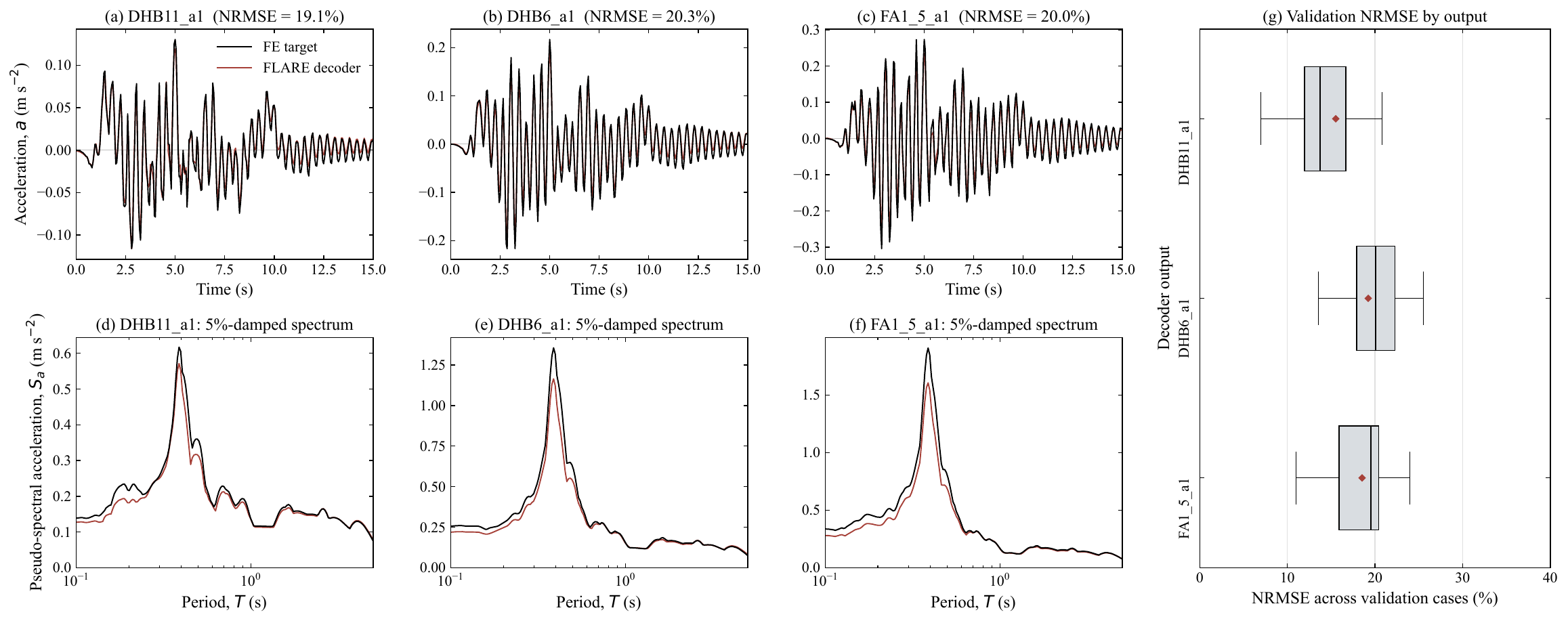}
\caption{Stage~2 validation: (a)--(c) sparse-response reconstructions, (d)--(f) corresponding $5\%$-damped pseudo-acceleration spectra, and (g) NRMSE distributions.}
\label{fig:lotung_stage2}
\end{figure}

Figures~\ref{fig:lotung_stage2}(a)--\ref{fig:lotung_stage2}(c) compare the finite-element responses and their reconstructions for representative validation Case~081 at DHB11, DHB6, and FA1-5. The reconstructed histories follow the phase, principal amplitudes, and decay of the target motions at all three locations, with NRMSE values of 19.1, 20.3, and 20.0\%, respectively. The corresponding $5\%$-damped pseudo-acceleration spectra in Figs.~\ref{fig:lotung_stage2}(d)--\ref{fig:lotung_stage2}(f) reproduce the period of the dominant spectral peak and the overall spectral shape, although the peak spectral amplitudes are underestimated to varying degrees. Figure~\ref{fig:lotung_stage2}(g) summarizes the results for Cases~081--100; each boxplot represents the NRMSE distribution of one output sensor over the 20 validation motions. DHB11 exhibits a lower median error, while the error distributions for DHB6 and FA1-5 remain similar to each other and within the same overall range. The consistent reconstruction of motions at the three elevations supports the use of a single sparse encoder to place the available sensor responses in the latent coordinates established by the dense-response model.

\subsubsection{Record-Based Calibration and Held-Out Prediction}
\label{sec:lotung_stage3}

The field records were divided by earthquake event, with LSST05, LSST06, LSST08, LSST09, LSST10, LSST12, LSST13, and LSST18 used for calibration; LSST11, LSST15, and LSST16 used for model selection; and LSST07, LSST14, and LSST17 retained for testing. The EW and NS components of each event were assigned to the same subset. Both directions used the source model and sparse encoder obtained from Stages~1 and 2, while the latent-dynamics coefficients were calibrated separately for each component. Only these coefficients were updated, and the test events were excluded from both coefficient adjustment and model selection.

\begin{figure}[!t]
\centering
\includegraphics[width=\textwidth]{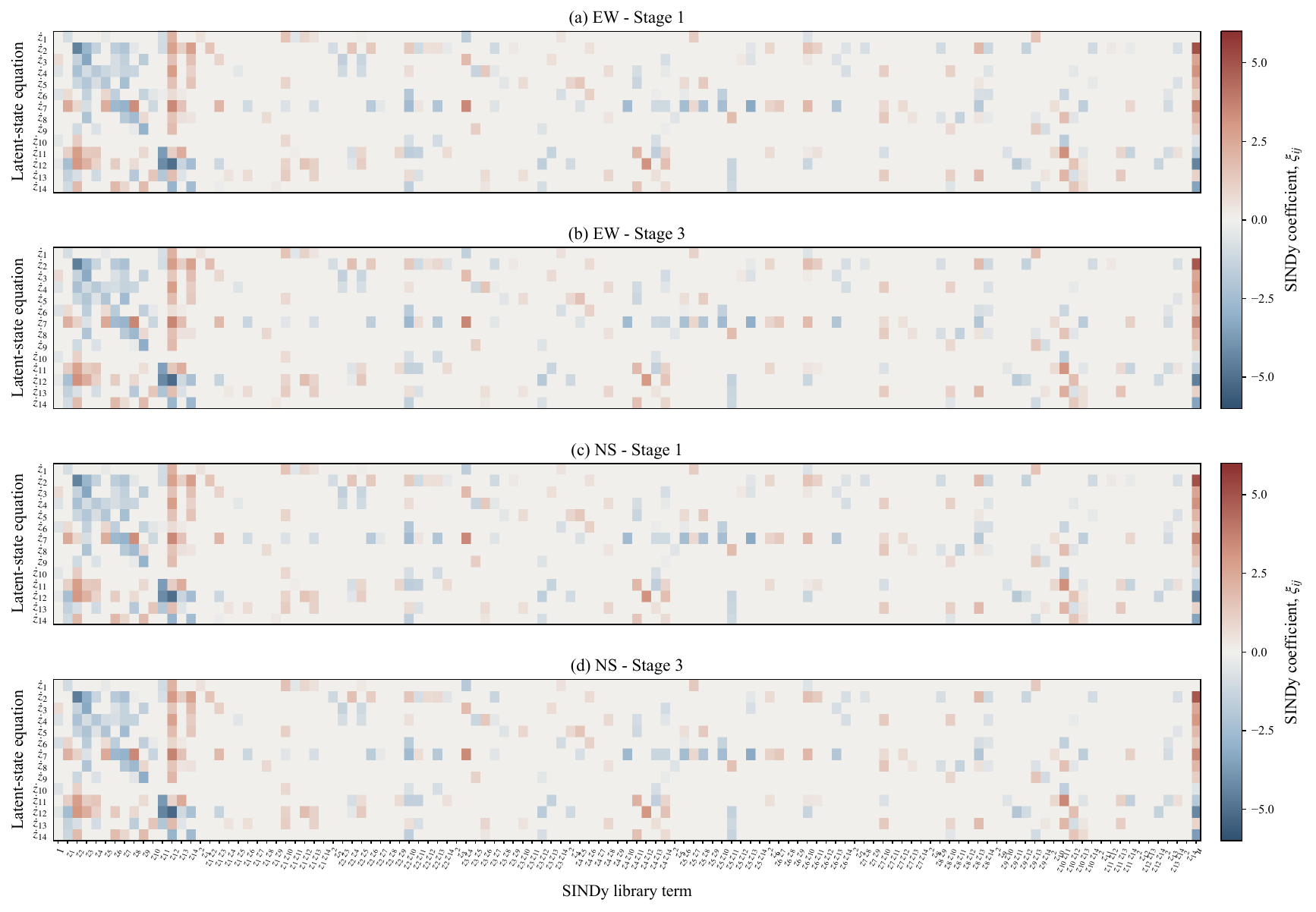}
\caption{Latent-dynamics coefficients for (a) EW at Stage~1, (b) EW at Stage~3, (c) NS at Stage~1, and (d) NS at Stage~3.}
\label{fig:lotung_coefficients}
\end{figure}

Figures~\ref{fig:lotung_coefficients}(a) and \ref{fig:lotung_coefficients}(b) compare the Stage~1 and Stage~3 coefficient matrices for the EW component, while Figs.~\ref{fig:lotung_coefficients}(c) and \ref{fig:lotung_coefficients}(d) provide the corresponding comparison for the NS component. The locations and signs of the dominant coefficients remain largely unchanged after calibration, indicating that both direction-specific models retain the principal latent dynamics learned from the numerical simulations. The measured records primarily modify the magnitudes of a limited subset of coefficients, and the modified entries differ between EW and NS. These differences arise from the independent calibration of the two measured components within the same latent coordinates and do not represent direction-dependent soil material parameters.

\begin{figure}[!t]
\centering
\includegraphics[width=\textwidth]{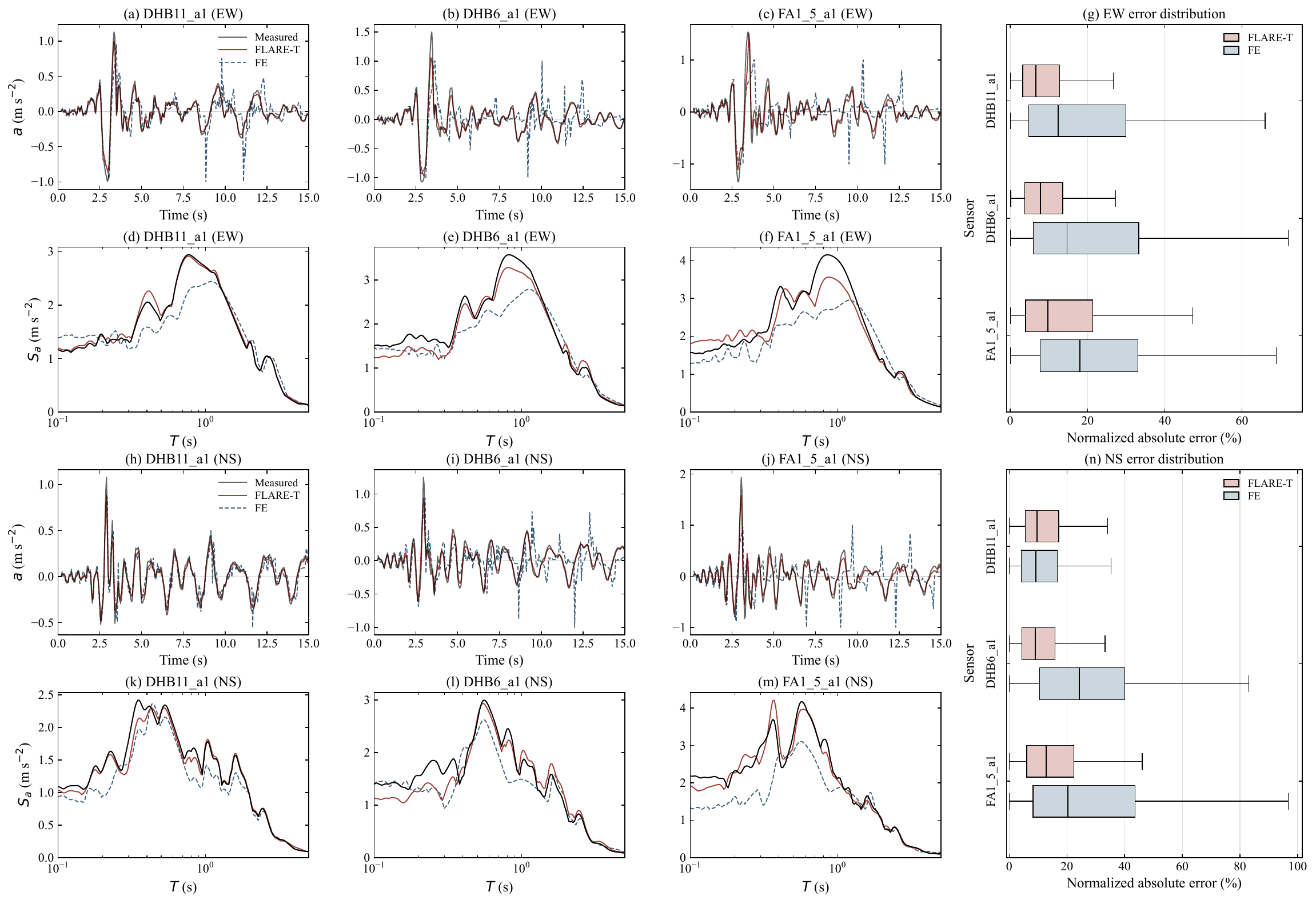}
\caption{LSST07 predictions: (a)--(g) EW and (h)--(n) NS time histories, response spectra, and error distributions.}
\label{fig:lotung_lsst07}
\end{figure}

\begin{figure}[!t]
\centering
\includegraphics[width=\textwidth]{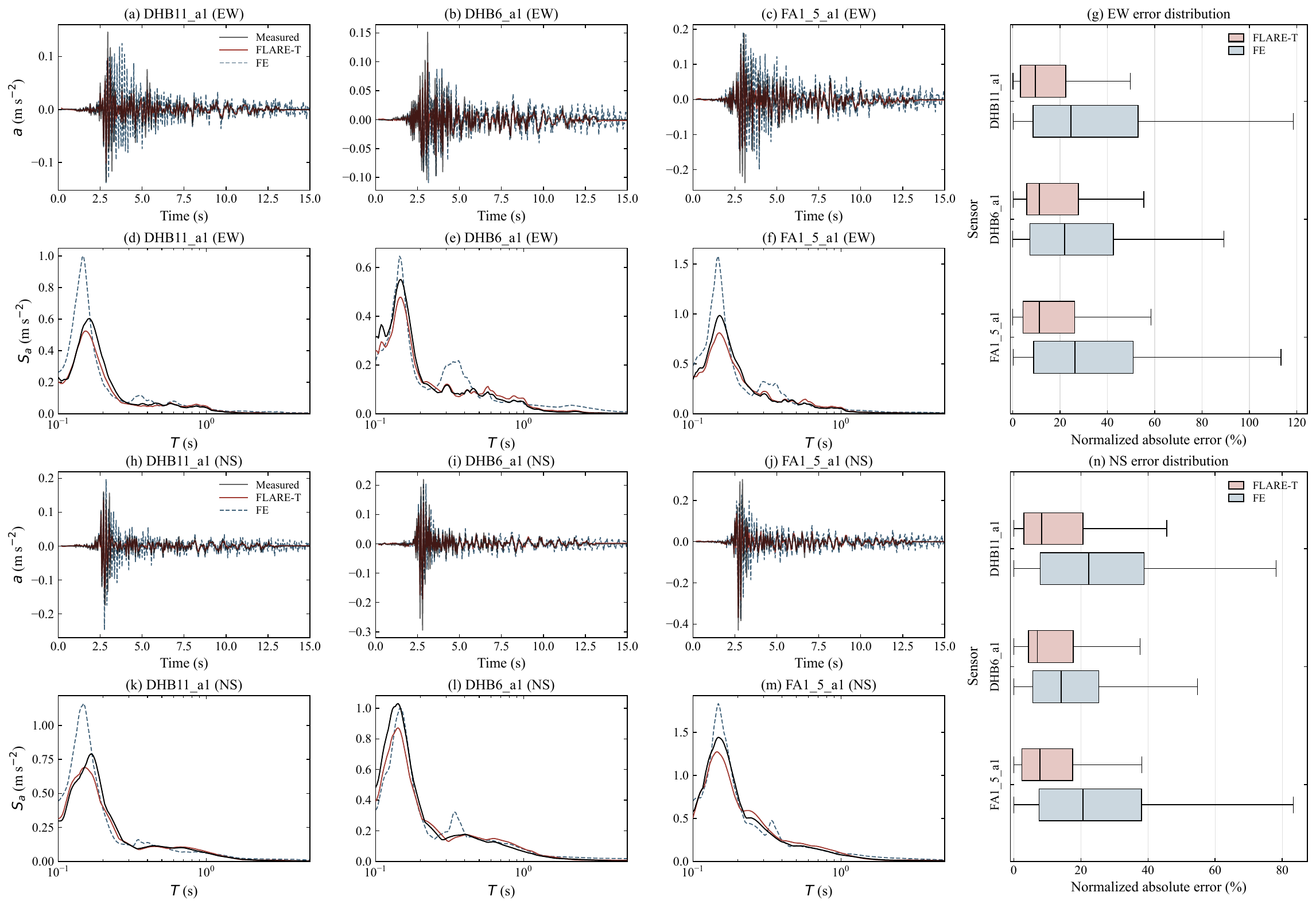}
\caption{LSST14 predictions: (a)--(g) EW and (h)--(n) NS time histories, response spectra, and error distributions.}
\label{fig:lotung_lsst14}
\end{figure}

\begin{figure}[!t]
\centering
\includegraphics[width=\textwidth]{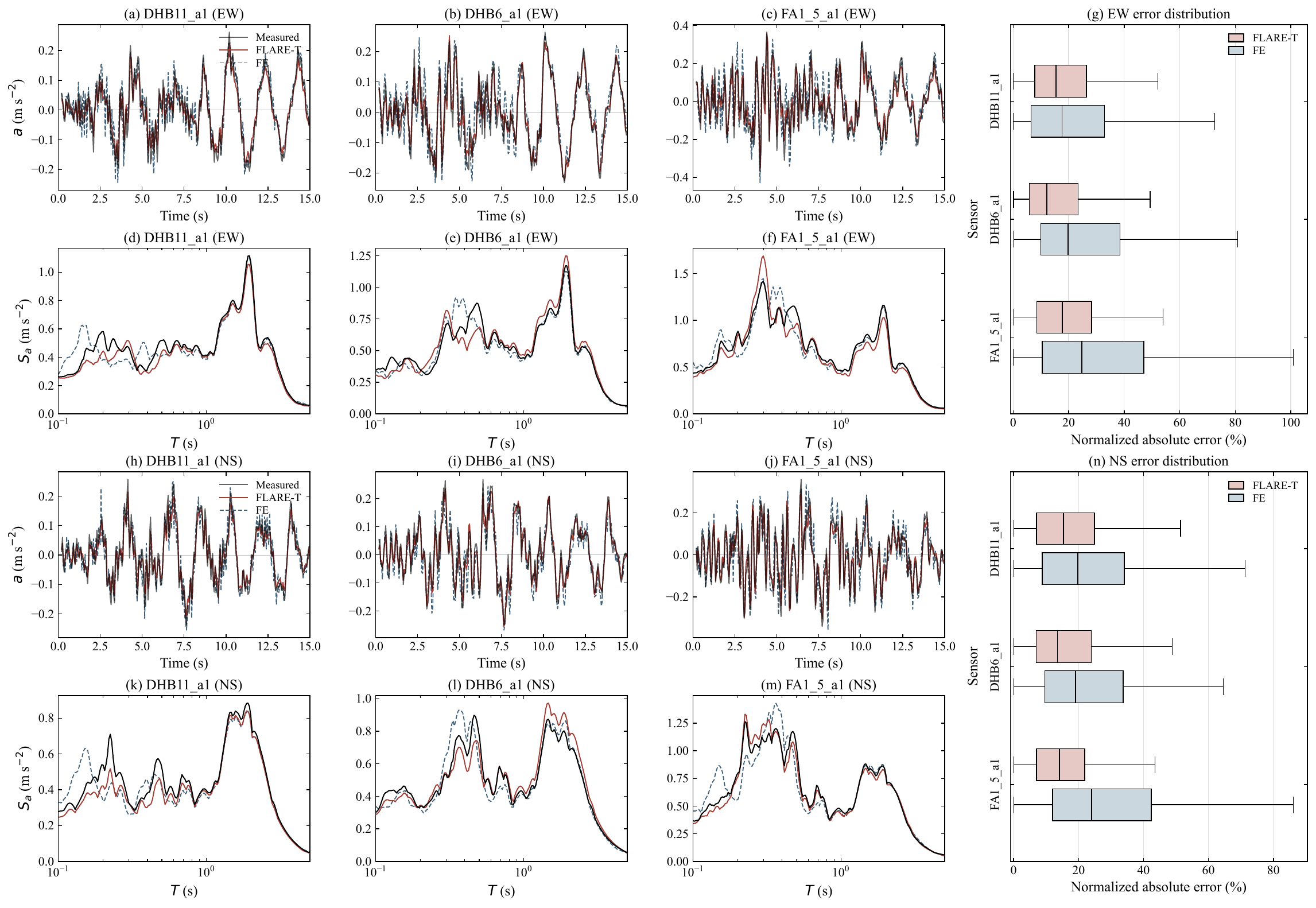}
\caption{LSST17 predictions: (a)--(g) EW and (h)--(n) NS time histories, response spectra, and error distributions.}
\label{fig:lotung_lsst17}
\end{figure}

Each held-out event was initialized using the same five-sample response window from DHB11, DHB6, and FA1-5. The complete DHB17 acceleration history was then supplied as the forcing input to predict the subsequent responses at the three upper sensors. FLARE-T predictions were compared with the measured records and the corresponding finite-element results before record-based correction. Timewise errors were evaluated using Eq.~\eqref{eq:timewise_error}, and the $5\%$-damped pseudo-acceleration spectra were evaluated over periods from 0.10 to 5.00~s. In Figs.~\ref{fig:lotung_lsst07}--\ref{fig:lotung_lsst17}, panels (g) and (n) show the EW and NS error distributions, respectively; each boxplot contains the timewise error samples for one sensor and one event component.

For LSST07, FLARE-T closely follows the phase and amplitudes of the principal EW and NS response cycles at DHB11 and DHB6 and substantially reduces the excessive late-time oscillations produced by the finite-element model. The improvement extends to FA1-5, where the timing of the principal surface pulses and the subsequent decay are reproduced more consistently in both directions. The predicted spectra also recover the dominant spectral bands at the three elevations more accurately than the finite-element results, particularly around the main intermediate-period peaks. The remaining surface discrepancies are concentrated near the largest acceleration pulses and the amplitudes of the dominant spectral peaks. Consistently lower error distributions in Figs.~\ref{fig:lotung_lsst07}(g) and \ref{fig:lotung_lsst07}(n) show that the improvement is present in both horizontal components.

LSST14 has smaller response amplitudes and a greater concentration of energy at short periods. FLARE-T reproduces the onset and decay of the measured oscillations in both directions, whereas the finite-element responses retain excessive high-frequency motion after the principal wave group. The short-period spectral peaks are also brought closer to the measured amplitudes at DHB11, DHB6, and FA1-5, although moderate underestimation remains near several peak ordinates. For every sensor, the FLARE-T error distributions lie below their finite-element counterparts in both Figs.~\ref{fig:lotung_lsst14}(g) and \ref{fig:lotung_lsst14}(n), demonstrating that the improvement is not confined to one horizontal component.

LSST17 contains multiple prominent wave groups distributed over a longer portion of the record. FLARE-T follows their phase evolution and depth-dependent amplitudes in both components, including the slower response cycles that dominate the latter part of the motion. The predicted spectra generally reproduce the locations and amplitudes of the principal peaks, while the finite-element results exhibit larger discrepancies in the short-period range and at the ground surface. The reductions in timewise error are comparable between EW and NS and occur at all three elevations, with slightly larger residual surface errors in the EW component. Local differences remain around individual peaks and intervals containing rapid changes in frequency content, but they do not obscure the principal response characteristics.

Across the three held-out events, FLARE-T provides more consistent predictions of response phase, amplitude evolution, and the principal spectral peaks than the original finite-element model. The timewise error distributions are reduced at all three sensor elevations for both horizontal components, covering motions with different amplitudes, frequency contents, and durations. These results demonstrate that a common numerical source model can be transferred to sparse field observations and subsequently calibrated into direction-specific predictive models. The following subsection examines how the initial parameterization of the numerical source model influences this transfer process.

\subsection{Influence of Prior Numerical-Model Calibration}
\label{sec:source_model_calibration}
A physics-based numerical source model is fundamental to FLARE-T because its simulated response fields provide the spatial information required to establish the latent response manifold and its input-driven dynamics. In engineering applications, however, the extent to which constitutive parameters can be calibrated before prediction depends on the available field observations. To examine the influence of this prior calibration, two numerical source models are considered for the Lotung site. The models share the same soil profile, shear-wave velocity distribution, SANISAND formulation, and numerical configuration, but use different values for selected constitutive parameters. This controlled comparison evaluates whether the record-based calibration stage can maintain consistent predictive performance for held-out events under different degrees of prior parameter calibration. The investigation therefore concerns the sensitivity of FLARE-T to specific parameter values within a common and physically representative numerical framework, for which the source model remains the basis of response learning and transfer.

The baseline source model adopted the initial SANISAND parameter set, with $m=0.01$, $h_0=7.05$, and $e_{\mathrm{ini}}=0.88$, whereas the precalibrated source model used $m=0.0085$, $h_0=6.00$, and $e_{\mathrm{ini}}=0.86$, as summarized in Table~\ref{tab:source_model_parameters}. These parameters affect the onset and accumulation of plastic deformation, the rate of hardening, and the initial soil state; their modification therefore produces meaningful differences in the simulated cyclic response. Because the elastic stiffness in SANISAND also depends on the void ratio, the layer-specific $G_0$ values were recalculated after changing $e_{\mathrm{ini}}$ so that the shear-wave velocity assigned to each layer remained equal to the profile presented in Fig.~\ref{fig:lotung_site}. This treatment allows the comparison to focus on differences in nonlinear constitutive response rather than differences in the initial wave-propagation velocity. The remaining SANISAND parameters and the numerical configuration were retained, and an independent simulated-response dataset was generated from each source model before applying the same three-stage FLARE-T procedure.
\begin{table}[t]
\caption{SANISAND parameters used in the two Lotung source models.}
\label{tab:source_model_parameters}
\centering
\begin{tabular}{lcc}
\hline
Parameter & Baseline source model & Precalibrated source model \\
\hline
$m$ & 0.0100 & 0.0085 \\
$h_0$ & 7.05 & 6.00 \\
$e_{\mathrm{ini}}$ & 0.88 & 0.86 \\
\hline
\end{tabular}
\end{table}

\begin{figure*}[t]
    \centering
    \includegraphics[width=0.9\textwidth]{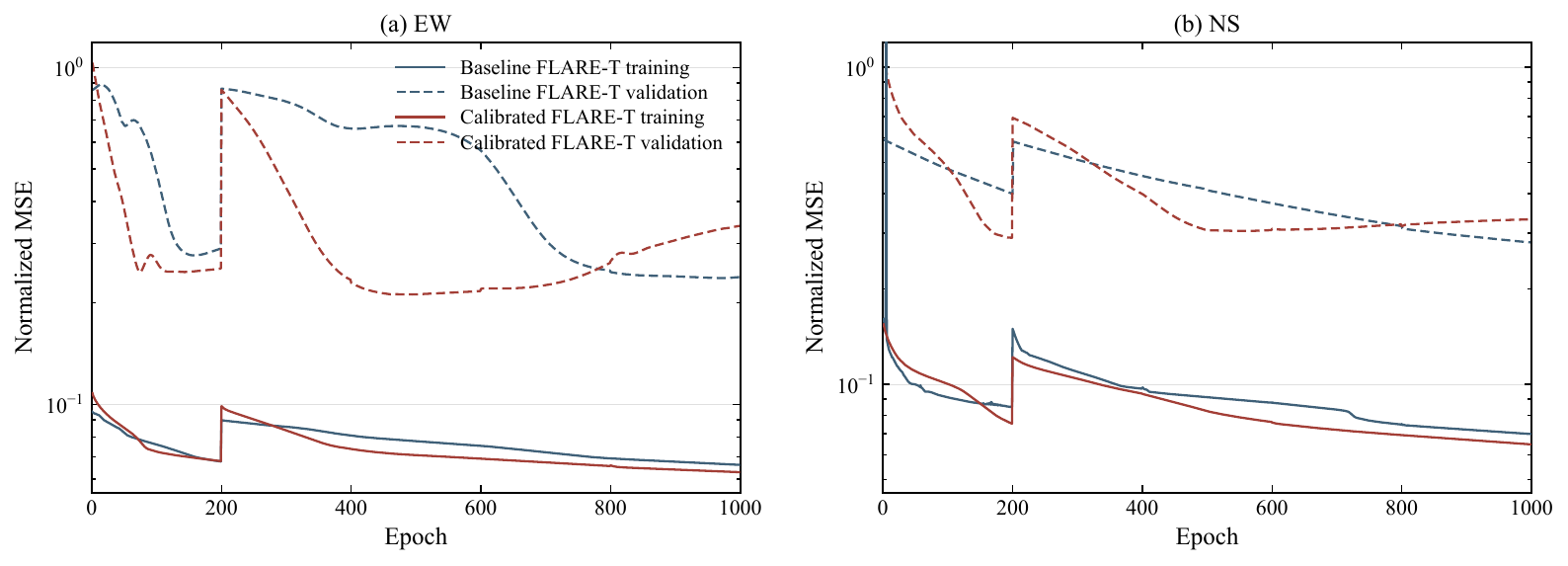}
    \caption{Third-stage optimization histories for the two source models: (a) EW and (b) NS.}
    \label{fig:source_model_training}
\end{figure*}

Figure~\ref{fig:source_model_training} compares the normalized training and validation MSE during third-stage calibration for the baseline and precalibrated source models, with the EW and NS results shown in Figs.~\ref{fig:source_model_training}(a) and \ref{fig:source_model_training}(b), respectively. For both horizontal components, the training errors decrease to comparable levels, while the validation errors remain within the same overall range despite differences in their convergence paths. The changes in loss at epoch 200 coincide with the first scheduled sequential thresholding operation, during which coefficients below the prescribed threshold are removed and the optimizer is reinitialized before calibration continues; the subsequent changes occur for the same reason. The bounded loss histories following these operations show that both source-model parameterizations can be calibrated using the available field records, although the initial parameterization affects the detailed optimization path and the validation error attained during training. For each source model and horizontal component, the model corresponding to the lowest validation error was retained for subsequent evaluation.

\begin{figure*}[ht]
    \centering
    \includegraphics[width=\textwidth]{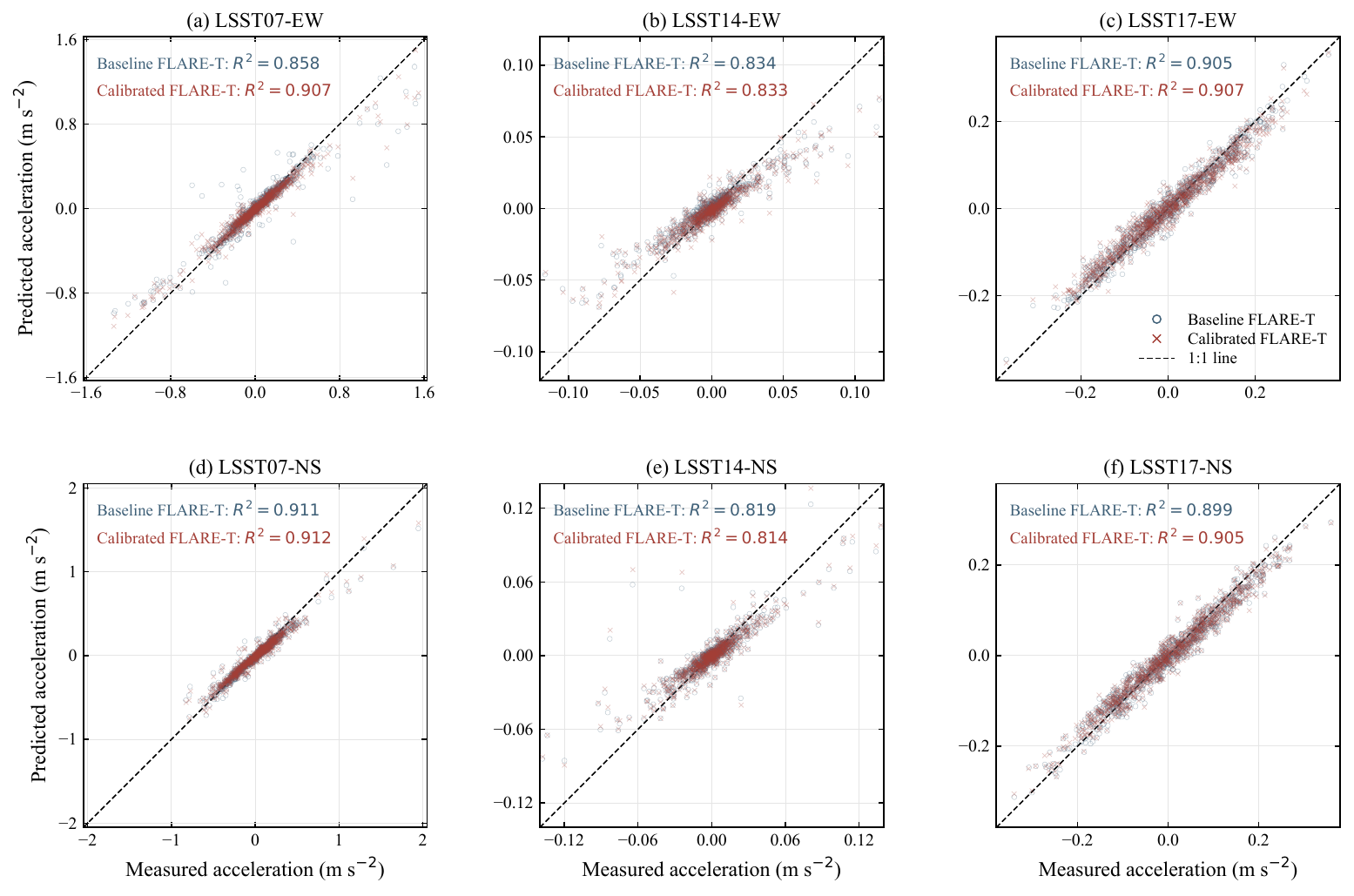}
    \caption{Measured and predicted accelerations for the held-out events: (a) LSST07-EW; (b) LSST14-EW; (c) LSST17-EW; (d) LSST07-NS; (e) LSST14-NS; and (f) LSST17-NS.}
    \label{fig:source_model_test}
\end{figure*}

Figure~\ref{fig:source_model_test} compares the held-out predictions obtained from the baseline and precalibrated source models. For each model, the third-stage coefficients selected from the validation events were used to predict LSST07, LSST14, and LSST17, none of which contributed to coefficient calibration or model selection. The latent state was initialized from the same five-sample response window, after which the complete DHB17 motion was applied as the external input. The predicted and measured accelerations at DHB11, DHB6, and FA1-5 were pooled over the prediction interval, so that each point in Fig.~\ref{fig:source_model_test} represents one measured--predicted acceleration pair at a particular sensor and time step. The agreement was quantified using the coefficient of determination,
\begin{equation}
R^2=1-\frac{\displaystyle\sum_{i=1}^{N}\left(\widehat{a}_i-a_i\right)^2}{\displaystyle\sum_{i=1}^{N}\left(a_i-\overline{a}\right)^2},
\label{eq:r_squared}
\end{equation}
where $N$ is the total number of pooled samples and $\overline{a}$ is their mean measured acceleration. The predictions from both source models are concentrated around the $1{:}1$ line, with $R^2$ ranging from 0.814 to 0.912 across the three events and two horizontal components. The clearest effect of prior numerical-model calibration occurs for LSST07-EW, for which $R^2$ increases from 0.858 to 0.907. Only small changes are observed for LSST07-NS and both components of LSST17, while the differences for LSST14 are also minor, with the baseline-source predictions giving slightly higher $R^2$. These results show that prior calibration can improve prediction for individual events, while third-stage calibration using field records produces broadly consistent held-out predictions from both numerical source models within the parameter range considered.

Taken together, the results of this chapter establish the complementary roles of numerical simulation and physical observations within FLARE-T. Across the centrifuge and Lotung applications, the dense numerical responses provided the depth-dependent response patterns and their input-driven evolution, the sparse-response transfer connected this information to the available sensor configuration, and the recorded motions corrected the transferred dynamics for prediction of held-out events. The resulting models reproduced multi-depth acceleration histories and the principal features of the corresponding $5\%$-damped pseudoacceleration spectra under both controlled experimental and field conditions. The comparison of the two Lotung source models further showed that the numerical model remains essential for providing a physically based response representation, whereas the final predictions depend less strongly on the precise prior calibration of the selected constitutive parameters. Better prior calibration can reduce the discrepancy presented to the transfer procedure and improve predictions for particular events, as observed for LSST07-EW, but comparable performance for the remaining events and components indicates that the record-based correction can compensate for a substantial portion of the initial parameter-related error. FLARE-T therefore provides a practical means of retaining the physical information supplied by a site-response model while reducing the level of prior parameter calibration required to obtain reliable predictions for subsequent earthquake events.

\section{Conclusion}
This study proposed the Transfer-Enabled Forced Latent Autoencoder for Response Equations (FLARE-T) to improve site-response predictions by correcting discrepancies between numerical simulations and measured motions using limited site-specific records. The soil profile is represented as an input--output system, with base acceleration as its input and accelerations at multiple depths as its outputs. FLARE-T transfers the depth-dependent response representation learned from dense numerical simulations to the available sensor configuration and subsequently calibrates the latent dynamics using recorded motions. The method was evaluated using a layered-soil centrifuge test and the Lotung field vertical array, with predictive performance assessed on separate test sets in terms of multi-depth acceleration histories and $5\%$-damped pseudoacceleration response spectra.

The test-set results showed that FLARE-T consistently provided more accurate site-response predictions than the original finite-element models in both applications. For the centrifuge test, the improvement was maintained across all instrumented depths and over the range of shaking intensities considered. For the Lotung site, lower prediction errors were obtained at all evaluated depths, for both horizontal components, and for motions with different amplitudes, frequency contents, and durations. In both cases, FLARE-T more accurately reproduced the response phase, amplitude evolution, decay, and principal spectral characteristics. The agreement across depths and test events further supports the ability of each calibrated model to reproduce coordinated responses at multiple locations through a shared low-dimensional latent state driven by the prescribed base motion. These results demonstrate that the response information transferred from numerical simulations can be effectively corrected using limited records to improve the prediction of test-set motions under both experimental and field conditions.

The comparison between the two Lotung source models further clarifies the respective roles of numerical simulation and field observation in FLARE-T. Despite their different parameter settings, the two Lotung source models produced broadly comparable predictions for the test set. The generally similar performance indicates that FLARE-T reduces the dependence of the final prediction on the precise prior calibration of the selected constitutive parameters, which is particularly valuable when the available site data are insufficient for exhaustive numerical-model calibration. This result does not diminish the importance of the initial numerical model. Its essential role is to provide a physically reasonable coordinate system for the site response by describing the principal patterns of wave propagation and the coordinated variation of acceleration across depth. This role is directly connected to the low-dimensional response-manifold assumption underlying FLARE-T: the soil profile is represented as a forced input--output system, in which base acceleration drives a low-dimensional latent state that describes the coordinated acceleration response across depth. From this perspective, the comparable predictions obtained from the two parameterizations suggest that both source models captured response patterns suitable for prediction, despite differences in their simulated evolution. Limited records could therefore correct much of the prediction discrepancy within the response representations supplied by these models. This distinction between representing the dominant response patterns and predicting their evolution explains how FLARE-T can reduce the demand for precise prior calibration while retaining the physical information provided by the numerical model. By addressing the sparse instrumentation and small event samples commonly encountered in engineering practice, the framework provides a practical route for converting existing numerical site models into observation-corrected predictive models for subsequent earthquake events.

\section*{Data and Code Availability}
Data and code generated or analysed during this study are available from the corresponding author upon reasonable request.

\section*{Declarations}
The authors have no conflicts of interest to declare that are relevant to the content of this article.

\section*{Acknowledgements}
This work is under the support of National Natural Science Foundation of China (Grant Numbers U2539204 and 52192675).

\bibliographystyle{ascelike-new}
\bibliography{references}

\end{document}